\documentclass{article}

\usepackage[preprint]{neurips_2025}

\usepackage[utf8]{inputenc} % allow utf-8 input
\usepackage[T1]{fontenc}    % use 8-bit T1 fonts
\usepackage{hyperref}       % hyperlinks
\usepackage{url}            % simple URL typesetting
\usepackage{booktabs}       % professional-quality tables
\usepackage{amsfonts}       % blackboard math symbols
\usepackage{nicefrac}       % compact symbols for 1/2, etc.
\usepackage{microtype}      % microtypography
\usepackage{xcolor}         % colors
\usepackage[pdftex]{graphicx}
\usepackage{subcaption}
\usepackage{float}
\usepackage{multirow}
\usepackage{amsmath}
\usepackage{algorithm}
\usepackage{algorithmic}

\usepackage{amsthm}
\newtheorem{theorem}{Theorem}

\title{InstCache: A Predictive Cache for LLM Serving}

\author{%
  Longwei Zou \thanks{* Equal contribution} \\
  \texttt{zoulw22@mails.tsinghua.edu.cn} \\
  \And
  Yan Liu \footnotemark[1] \\
  \texttt{liuyan22@mails.tsinghua.edu.cn} \\
  \And
  Jiamu Kang \\
  \texttt{kangjiamu1213@gmail.com} \\
  \And
  Tingfeng Liu \\
  \texttt{ltf23@mails.tsinghua.edu.cn} \\
  \And
  Jiangang Kong \\
  \texttt{kongjiangang@didiglobal.com} \\
  \And
  Yangdong Deng \\
  \texttt{dengyd@tsinghua.edu.cn} \\
}

\begin{document}

\maketitle

\setcounter{footnote}{0}

\begin{abstract}
The revolutionary capabilities of Large Language Models (LLMs) are attracting rapidly growing popularity and leading to soaring user requests to inference serving systems. Caching techniques, which leverage data reuse to reduce computation, offer opportunities to optimize the performance of LLM inference engines. On the one hand, the low-level key-value (KV) cache working at the token level is widely adopted, albeit it incurs significant overhead as request volume grows. On the other hand, instruction-level caching, which stores full instruction-response pairs, is expected to play an increasingly crucial role. However, the high variability in the content and length of instructions make it rare for identical instructions to recur within a short time window, presenting challenges for effective caching instruction-response pairs. To address this challenge, we propose InstCache, a predictive caching mechanism for LLM serving systems. Leveraging the capability of LLMs, we can effectively reorder the representation space of instruction texts and develop a sufficient level of spatial locality. Such spatial locality enables us to predict potential instructions located in a compact region in the space, resulting in an effective caching system at runtime. Experimental results demonstrate that InstCache achieves a 2.3× higher hit rate compared to the upper bound of traditional caching mechanisms on WildChat dataset and reduces the time per output token of vLLM by up to 42.0\% and 50.0\% on LMSys and Moss datasets, respectively.\footnote{Code is available at https://anonymous.4open.science/r/InstCache-1244/}

\end{abstract}

\section{Introduction}
\label{introduction}

With the rapid growth of demands from a fast increasing number of users and applications, the design of efficient serving system has become a critical challenge. For example, ChatGPT\citep{openai} has reached 400 million weekly active users in February 2025 and is expected to exceed one billion by the end of this year.\footnote{https://www.demandsage.com/chatgpt-statistics/}

Caching, as an effective method trading storage for computation, has long been a cornerstone of optimizing computing systems and received substantial attention in enhancing the performance of LLM serving systems. At a lower level, the Key–Value (KV) Cache accelerates next-token prediction within the conversation. At a higher level, instruction–response cache stores complete answers to frequently asked queries. As LLM requirements scale, instruction-response cache is expected to play an even more important role to enable computation savings. The unique characteristics of LLM instructions, however, hinders the efficiency of adopting such cache structures in real-world LLM inference serving systems.

\begin{figure}
    \centering
    \subcaptionbox{Token Length Distribution\label{fig:length_dist}}
    {\includegraphics[width=0.44\linewidth]{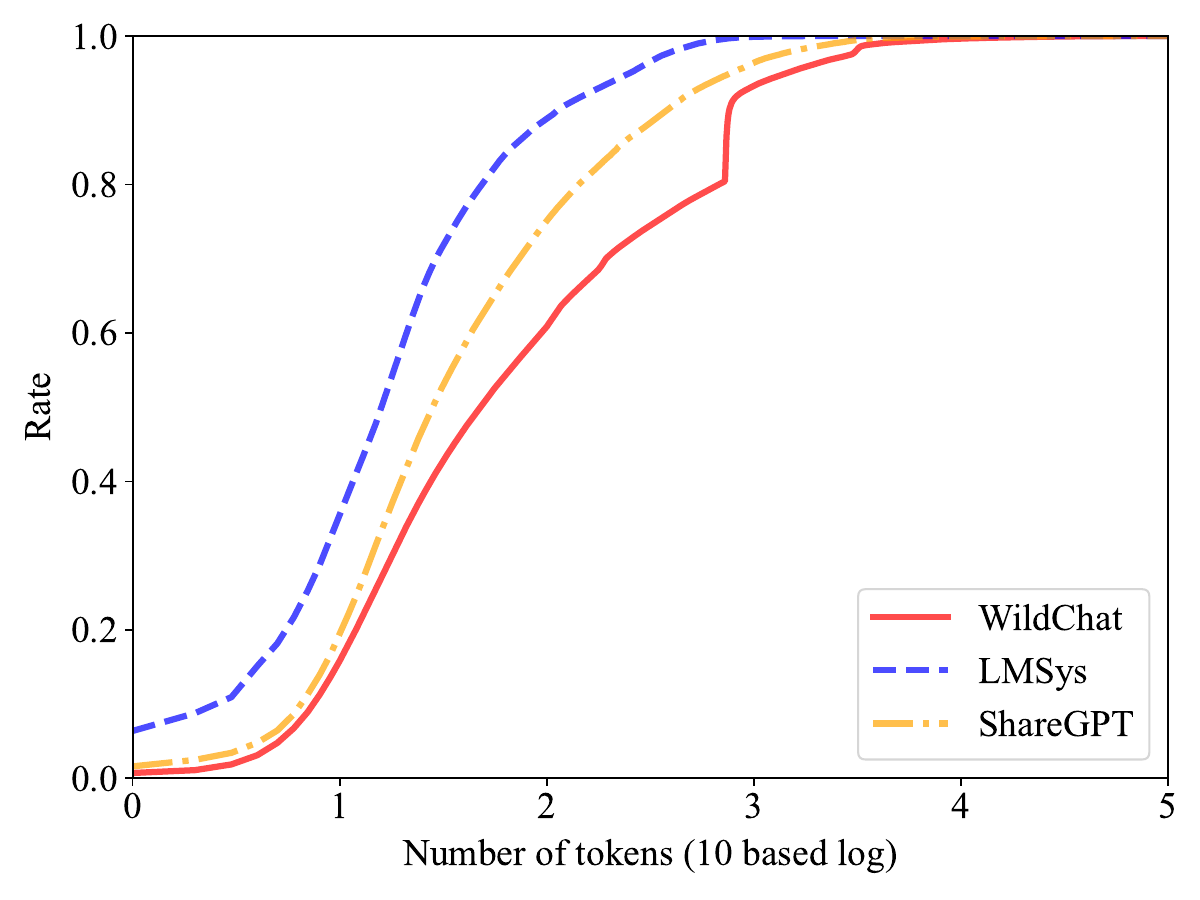}}
    \hspace{0.4cm}
    \subcaptionbox{Frequency of WildChat Dataset\label{fig:freq_wildchat}}
    {\includegraphics[width=0.44\linewidth]{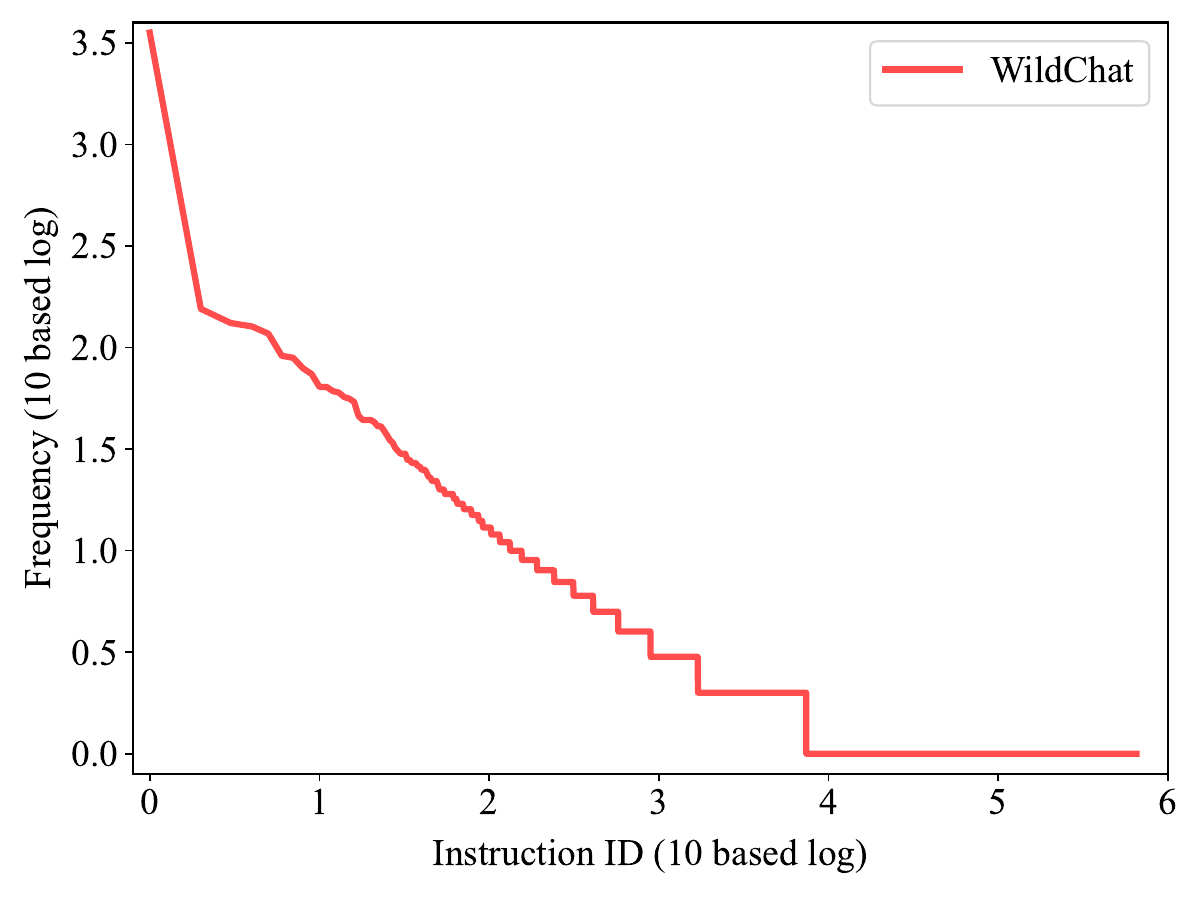}}
    \caption{Subfigure (a) presents the token length distribution of first-turn instructions across three real-world conversation datasets. Subfigure (b) shows the frequency of instructions in the WildChat dataset, sorted in descending order. The analysis reveals that most instructions are relatively long and exhibit minimal repetition across requests.}
    \label{fig:dataset_stat}
    \vspace{-1cm}
\end{figure}

Unlike web search engines, where query-result caching is effective, LLM queries are typically much longer and  more diverse. The WildChat\citep{wildchat} dataset, a real-world LLM serving dataset with one million conversations in the span of one year, illustrates such a trend. Figure \ref{fig:length_dist} presents the length distribution of first-turn instructions of WildChat, which shows significantly long input lengths. Statistics from other real-world dialogue datasets, like LMSys\citep{lmsys}, ShareGPT\citep{sharegpt}, also support the above observation. Prior research\citep{querylocality} indicates that most web search queries contain words fewer than five. The vast input space resulting from longer instructions leads to an exponential explosion in the number of possible instructions, and exact repetition is thus rare. Figure \ref{fig:freq_wildchat} demonstrates the frequency of first-turn instructions in the WildChat dataset. To avoid noise from meaningless and testing inputs like "hello" or "test", we filtered out instructions whose responses were shorter than 32 tokens, i.e., approximately two sentences. We also removed identical instructions submitted from the same hashed IP address, as these are typically test submissions. We do not profile other datasets because they lack sufficient information like hashed IP address to effectively filter out noise instructions. After filtering, we found that the two most frequent instructions appeared 3,551 and 155 times, respectively, and only 3.7\% of instructions were repeated. These results highlight the extremely sparse temporal locality in LLM serving instructions, posing a significant challenge for higher-level caching.

To address the sparsity problem, semantic caching\citep{GPTCache} has been proposed, matching instructions based on sentence embedding similarity. However, our preliminary experiments show that even with a high similarity threshold, semantic caches still frequently result in mismatches. Using DeepSeek V3\citep{deepseek_v3} to evaluate cache hit quality on the WildChat dataset, we found that approximately 30.0\% of the hit instructions are mismatched, which is considerable. For instance, GPTCache may mismatch instructions such as "how to create blog post" and "how can you help me on creating my blog site". The details of the evaluation is presented in Appendix \ref{GPTCache}. Such a high mismatching rate makes semantic caching unsuitable for real-world deployment. Furthermore, semantic caches inevitably introduce additional latency from sentence embedding and raise the concern for data leakage\citep{dataleakage}.

To address this challenge, we propose to leverage the intrinsic semantic capability of LLMs to reorder the text space and develop a sufficient level of spatial locality. Specifically, we show that each LLM is related to a text space. After fitting into instructions, the LLM can reorder the text space and gather potential instructions into a close neighborhood based on its semantic capability, thereby inducing spatial locality. Such a reordering enables the prediction of high-likelihood instructions and thus allows the construction of efficient caching systems. Conceptually, the LLM acts analogously to serving as a memory address remapping mechanism, facilitating continuous memory access and effective prefetching by transforming sparse accesses into a denser access pattern. By exploiting this emergent spatial locality, our method circumvents the limitations of traditional caches and offers a practical pathway to efficient cache instruction-response for large-scale LLM serving systems.

Our experimental results show that InstCache can achieve a hit rate that is 2.3x higher than the upper bound of that of traditional caches on the WildChat dataset. On the de-duplicated LMSys and Moss datasets, InstCache achieves hit rates of 20.1\%, and 23.4\%, respectively, demonstrating its predictive capability. When integrated with vLLM, InstCache reduce the time per output token by up to 42.0\% and 50.0\% on the LMSys and Moss datasets. Furthermore, we observe that the cache hit rate steadily improves as the volume of observed data increases, revealing the potential of InstCache.

\section{Related Work}

LLM serving systems receive an overwhelming number of requests daily. Caching systems, including KV cache and semantic cache, have been proposed to efficiently process these requests. Moreover, considering the similar role, we also introduce the query-result cache in web search engine area here.

\paragraph{KV Cache} Key-Value Cache (KV Cache) reuses the key-value states of previous tokens to accelerate the generation of subsequent tokens, serving at a lower level of LLM serving systems. When serving LLMs, key-value states from different conversations will quickly exhaust memories of GPU and CPU. To address this challenge, several works have been proposed. PageAttention\citep{vllm} suggests to manage the KV Cache by virtual memory management, minimizing the overhead of memory allocation. Other works, such as H\textsubscript{2}O\citep{H2O} and AttentionSink\citep{attentionsink}, focus on discarding key-value states of less important tokens. To further minimize storage requirement, SGLang\citep{sglang} and ChunkAttention\citep{chunkattention} propose to share common prefix KV Caches. Multi-Head Latent Attention\citep{mla} compresses the key-value states by low rank projection, reducing the storage cost. Additionally, AttentionStore\citep{attentionstore} suggests to store the KV Cache in multi-level storage systems. Since KV-Cache focuses on lower level decoding operations, it's orthogonal to InstCache proposed in the work.

\paragraph{Semantic Cache} In the Semantic Cache, as proposed in GPTCache\citep{GPTCache}, each instruction is embedded as a vector, which is stored in a database along with the instruction and response. Semantic cache retrieves response for an incoming instruction by searching for the most similar one in the database and reuses the answer of the previously query upon a hit, serving at a higher level of LLM serving systems. Following the same idea, MeanCache\citep{meancache} introduces a user-centric semantic caching system that preserves privacy of users. MeanCache also employs federal learning to build different embedding models locally to enhance the performance. There are also methods\citep{scalm, contextcache} aiming to improve the embedding quality through preprocessing instructions. Notably, different from previous methods evaluated on LLM-synthetic datasets\citep{GPTCache, input_gen}, SCALM\citep{scalm} proposes to evaluate the effectiveness of semantic cache on real-world datasets, such as LMSys dataset. While semantic cache mitigates the sparsity problem of text space by approximate matching, it's hard to deploy semantic cache in real-world LLM serving systems due to its high mismatching rate, as explained in Appendix \ref{GPTCache}. Moreover, semantic cache inevitably introduces additional embedding and similar vector searching latency, which is also considerable.

\paragraph{Search Engine Cache} Caching systems for search engine share many characteristics with LLM Cache. As search engine process keyword queries and return links to webpages that meet users' needs, the web cache saves queries and corresponding results to provide potential reusing opportunities. In early works\citep{querycache, querylocality}, researchers found that web search queries  usually contain fewer than five words and have a highly repetitive possibility. They propose to store the results of queries with different policies like LRU, FBR and SLRU. After that, various works have been proposed to provide more efficient caching system for web searching\citep{sigir2001, www2003, spire2003, www2005, tois2006, www2008, hpdc2010} from aspects of better policies or more intermediate result like inverted lists. However, due to much longer length, the repetitive possibility of instructions in LLM serving systems drop dramatically. Therefore, using caching systems from web search engine will provide minimal hit rate in the LLM serving system.

\section{Method}
\label{sec:instcache}

\begin{figure}[t]
    \centering
    \includegraphics[width=0.9\linewidth]{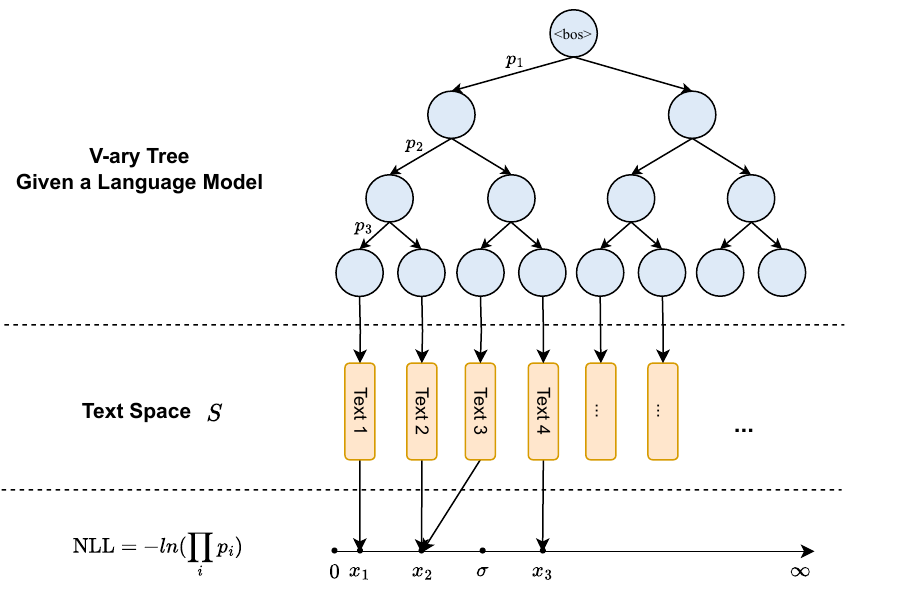}
    \caption{Text space can be represented as paths in a V-ary tree of depth $L$, constructed by a language model with a vocabulary size of $V$ and a maximum token length of $L$. In this V-ary tree, each node corresponds to a token, and each edge represents the probability of generating that token given the preceding context. The child nodes of any given node are ordered by their probabilities. Each path from the root to a leaf defines a unique text, collectively forming a sequential text space. Given the NLL of each path, these text sequences can be projected onto a one-dimensional space. For simplicity, the figure demonstrates a binary tree and assumes that each node’s children follow an identical probability distribution. Under this assumption, text2 and text3 share the same NLL.}
    \label{fig:text_space}
\end{figure}

Given a language model with a vocabulary size of $V$ and a maximum token length $L$, the entire text space can be represented as paths in a V-ary tree of depth $L$. In the V-ary tree defined by a language model, each node corresponds to a token, and each edge represents the probability of generating the next token given the preceding context. The child nodes of each node are ordered according to their respective generation probabilities. As illustrated in Figure \ref{fig:text_space}, each path from the root to a leaf in the tree corresponds to a unique text. Notably, each path should begin with \texttt{<bos>} token and end with \texttt{<eos>} token. Text ending with \texttt{<eos>} will not be expanded further. For simplicity, all paths shown in Figure \ref{fig:text_space} have the same length.

Given the text space, we use the Negative Log-Likelihood (NLL), defined as $\text{NLL}=-ln(\prod_{i=1}^{n}{p(t_i|t_1,t_2,...,t_{i-1})})$ to represent the position of a text within the space, analogous to a memory address. Given a specific language model, this induces a distribution of texts over a one-dimensional space. 

When a language model is trained on a specific class of texts, such as user instructions, texts with similar structures and meanings tend to have lower NLL values and are thus concentrated near the "head" of the space. Such a process induces a form of spatial locality, allowing the retrieval of the set of potential instructions by selecting a contiguous region near the beginning of the space.

Formally, given a threshold $\sigma$, we define the InstCache as $C=\{\langle s, r \rangle: \text{NLL}(s) \leq \sigma, s \in S\}$, where $s$ and $r$ denote an instruction and its corresponding response, and $S$ represents the text space. The cache therefore includes all instructions with their NLL falling below the threshold, i.e., the most likely ones.

This formulation allows us to analytically predict the hit rate with the threshold $\sigma$. Let the $\mathcal{S}$ denote a random variable of instruction and $\mathcal{N}=\text{NLL}(\mathcal{S})$ its associated $\text{NLL}$. The expected hit rate of InstCache is then given by:

\begin{align}
    \text{Hit}\_\text{Rate} &= P(\mathcal{S} \in C) \notag \\
              &= P(\mathcal{N} \leq \sigma) \notag \\
              &= F_{\mathcal{N}}(\sigma)
    \label{eq:hit_rate}
\end{align}

where $F_{\mathcal{N}}$ is the cumulative distribution function (CDF) of $\mathcal{N}$. By measuring this CDF on a validation set drawn from recently observed instructions, we can estimate the expected hit rate for a given cache threshold $\sigma$.

In preliminary experiments, we observe that the next-token probability distribution at each node tends to follow a power-law pattern. We further prove that the number of reachable leaf nodes in the text space grows as an exponential function depending on $\sigma$ when assuming a power-law distribution of the next-token. The formal theorem is stated as follows:

\begin{theorem}[Text Count Estimation]
\label{thm:text_count_estimation}

Given a language model with a vocabulary size $V$, where at each step of next-token generation, the model assigns a probability $P(t_i) = \beta i^{-\alpha}$ to each token $t_i$, with $i$ denoting the rank of the token. Let $L$ be the length of generated texts, and the generation starts from a \textless bos\textgreater{} token. Then the total number $N$ of distinct texts of length $L$ that can be generated is given by: $N \approx {e^{\frac{\sigma}{\alpha}} \cdot \left(\frac{\sigma}{\alpha}\right)^{L-1}}/{(L-1)!}$

\end{theorem}

Details of the power-law pattern observation and proof are provided in Appendix \ref{sec:theorem_prove}. The power-law distribution parameters $\alpha$ and $\beta$ can be estimated based on actual next-token generation statistics. In Section \ref{sec:caching_results}, we evaluate the accuracy of our hit rate and cache size prediction model. Since the distribution of instructions evolves gradually over time, we also examine this effect in Section \ref{sec:distribution_shift} and demonstrate that the most popular instructions tend to remain stable and do not change frequently.

\section{Implementation}
\label{sec:implementation}

\subsection{Cache Pre-population}

Before pre-populating InstCache, we first train a LLM using observed instructions, padding them with \texttt{<bos>} and \texttt{<eos>} tokens. The pre-population LLM is typically significantly less complex than the model to be served. Afterward, we traverse the V-ary tree step by step to pre-populate InstCache with instructions whose $\text{NLL}$ values are below the threshold $\sigma$ and that end with the \texttt{<eos>} token.

\begin{algorithm}[h]
   \caption{Instruction Pre-population}
   \label{alg:tree_search}
\begin{algorithmic}
   \STATE {\bfseries Input:} a fine-tuned LLM, Max NLL $\sigma$, Max Length $L$
   \STATE {\bfseries Output:} Pre-populated instructions $C$
   \STATE Initialize an empty dequeue $Q$ and empty list $C$
   \STATE Initialize root $r$ with \texttt{<bos>} token
   \STATE Enqueue $r$ into $Q$
   \WHILE{$Q$ is not empty}
   \STATE Dequeue a node $u$ from the head of $Q$
   \STATE Generate next tokens $T$ for $u$ with LLM 
   \STATE // Suppose next tokens are sorted in ascending order of probability
   \FOR{each token $t$ of $T$}
   \IF{Length from $r \to u$ $<$ $L$ and NLL from $r \to t$ $\leq$ $\sigma$}
   \IF{$t$ is \texttt{<eos>}}
   \STATE Decode token sequence from $r$ to $t$ and append to $C$
   \ELSE
   \STATE Add $t$ as a child of $u$
   \STATE Enqueue children node $t$ into the head of $Q$
   \ENDIF
   \ENDIF
   \ENDFOR
   \ENDWHILE
\end{algorithmic}
\end{algorithm}

As shown in Algorithm \ref{alg:tree_search}, the tree is traversed in a depth-first manner. At each step, the model generates tokens for the current node. Due to the cumulative nature of NLL, we can prune child nodes with their NLLs exceeding the threshold, thus avoiding unnecessary computations. To minimize redundant recomputation, the key-value states associated with each node are preserved during traversal until all its descendants have been explored. The depth-first strategy enables promptly releasing these key-value states, thereby reducing GPU memory consumption.

Due to the preserved KV cache, each node generation corresponds to a decoding step in LLM inference. For clarity, Algorithm \ref{alg:tree_search} illustrates the method using a batch size of 1. However, in practice, we adopt large-batch decoding to fully leverage the computing power of GPUs. In Section \ref{sec:prepopulation_cost}, we demonstrate the cost of pre-population and show that pre-populating 10,000 instructions using LLaMA3-8B on an A100 GPU takes only about 1 minute, which is not a significant burden for an enterprise.

% TODO: 上面这个具体时间数据需要检查

Response generation for pre-populated instructions is performed offline using serving systems such as vLLM\citep{vllm} and SGLang\citep{sglang}. These systems can exploit shared prefixes among instructions to further accelerate response generation.

Since the expansion of each node is independent, InstCache is inherently parallelizable. One practical strategy is to first explore the tree for a limited number of steps and then distribute the remaining leaf nodes across multiple worker processes. Section \ref{sec:prepopulation_cost} demonstrates that such a parallelization strategy significantly improves pre-population speed with minimal overhead.

\subsection{Cache Deployment}

To minimize the lookup latency of InstCache, we store pre-populated instructions and the respective responses in a hash table, which maps instructions to their corresponding responses. Due to its memory and CPU-intensive nature, InstCache is orthogonal to existing LLM serving systems and can be easily integrated with them. When a submitted instruction is found in the cache, the corresponding response is immediately returned to the user. In case of a cache miss, the instruction is forwarded to the LLM serving system for processing. The additional latency introduced by InstCache is negligible, as the lookup complexity is close to $O(1)$. By leveraging a distributed key-value database, InstCache is also well-suited for distributed deployment, which can further increase the hit rate by caching more instructions and responses in distributed or multi-level storage.

\section{Experiment}

\subsection{Experimental Setup}
\label{sec:exper_setup}

\paragraph{Datasets} 
We use WildChat\citep{wildchat}, LMSys\citep{lmsys}, ShareGPT\citep{sharegpt} and Moss\citep{moss} as our evaluation datasets. WildChat is a corpus of 1 million user-ChatGPT conversations, collected by offering free access to ChatGPT for online users in exchange for their affirmative. LMSys is obtained from the Vicuna demo and the Chatbot Arena website\citep{arena}, encompassing approximately 1 million conversations. ShareGPT consists of 100,000 realistic conversations shared by GPT users. Moss contains 1.1 million synthetic conversations generated by LLM paraphrase based on collected dialogues. Unless otherwise specified, following experiments are conducted on instructions with token lengths below 128 and focus on the first turn of each conversation. Since over half of conversations contain only a single turn\citep{wildchat} and the majority of first-turn instructions are under 128 tokens, this setting captures most of the computational workload in these datasets.

For these datasets, we examine the quality of user instructions and identified a significant number of repetitive and test prompts, such as "hi", "test", and "111". To reflect real-world application scenarios and mitigate the bias introduced by these instructions that can artificially boost cache hit rates, we perform de-duplication to filter the datasets. Specifically, for WildChat, leveraging the available IP information, we remove duplicate instructions originating from the same hashed IP address. For the remaining datasets, we eliminate all identical instructions. Furthermore, for all datasets, we filter out instructions whose ChatBot-responses are shorter than 32 tokens, as these are typically meaningless instructions and unlikely to make substantial throughput improvements even with cache hits.

\paragraph{Baselines} 

Since instructions exhibit minimal repetition after filtering, we compare InstCache against the theoretical upper bound of traditional exact matching cache. Specifically, we directly utilized 80\% of the data, which is consistent with the 80\% used for training the model, to construct the cache, and measured the hit rate on the remaining 20\%, which reflects the degree of repetition. Before computing the repetition rates, we lowercase all instructions, following common practice in caching systems. As a result, the deduplicated datasets have a very low repetition rates. Additionally, GPTCache employs semantic matching rather than exact matching, leading to a high rate of mismatches, as explained in Appendix \ref{GPTCache}. Such a high mismatching rate significantly limits the practicality of GPTCache in real-world deployments. For fairness, we exclude GPTCache from comparison with InstCache, which employs an exact matching mechanism.

\paragraph{Implementation Details}

For InstCache, we split each dataset into 80\% for fine-tuning the LLM, leaving 10\% for validation and 10\% for test set. Appendix \ref{sec:data_scale} presents an evaluation of the impact of varying training and test data ratios. Our preliminary experiments revealed that using a base LLM model for cache pre-population outperforms models that have undergone supervised finetuning (SFT) and reinforcement learning with human feedback (RLHF), likely due to the alignment tax problem\citep{alignment_tax}. To balance model capability and the cache pre-population speed, we use a relatively small LLM, i.e. LLaMA3-8B\citep{llama3}. In fact, we tested various models of similar sizes and observed comparable performance. We fine-tune all parameters of the model for 2 epochs with a batch size of 384, a maximum sequence length of 128 tokens, a weight decay of 0.1, and a learning rate of 2e-5 across all the datasets.
Our hardware platform comprises three A100 GPUs, each equipped with 80GB of memory. The pre-population experiment without distributed processing is conducted on a single GPU, while the distributed version utilizes all three GPUs. We use a batch size of 2048 for the cache pre-population. For the evaluations in Section \ref{sec:integration_w_sys}, we deploy Qwen2.5-32B-Instruct\citep{qwen2.5} model using the vLLM serving system on two A100 GPUs.

The cache constructed with Algorithm \ref{alg:tree_search} includes all predicted user instructions. To enable deployment in real-world application scenarios, each of these instructions must be populated with a corresponding response. This response generation step is decoupled from our core method. Depending on the desired service quality, different model sizes can be employed for this step. Alternatively, even closed-source model APIs could be utilized. In our experimental setup, we employed Qwen2.5-32B-Instruct to generate 5,000 sampled instructions from each cache and estimate the total cache size based on the average length of the generated responses. The estimated average response lengths are closely aligned with those in the original datasets. Details of the constructed caches are provided in Appendix \ref{sec:cache_details}

\subsection{Experimental Results}

\subsubsection{Caching Results}
\label{sec:caching_results}

\begin{figure}[h]
    \centering
    \subcaptionbox{WildChat}
    {\includegraphics[width=0.262\linewidth]{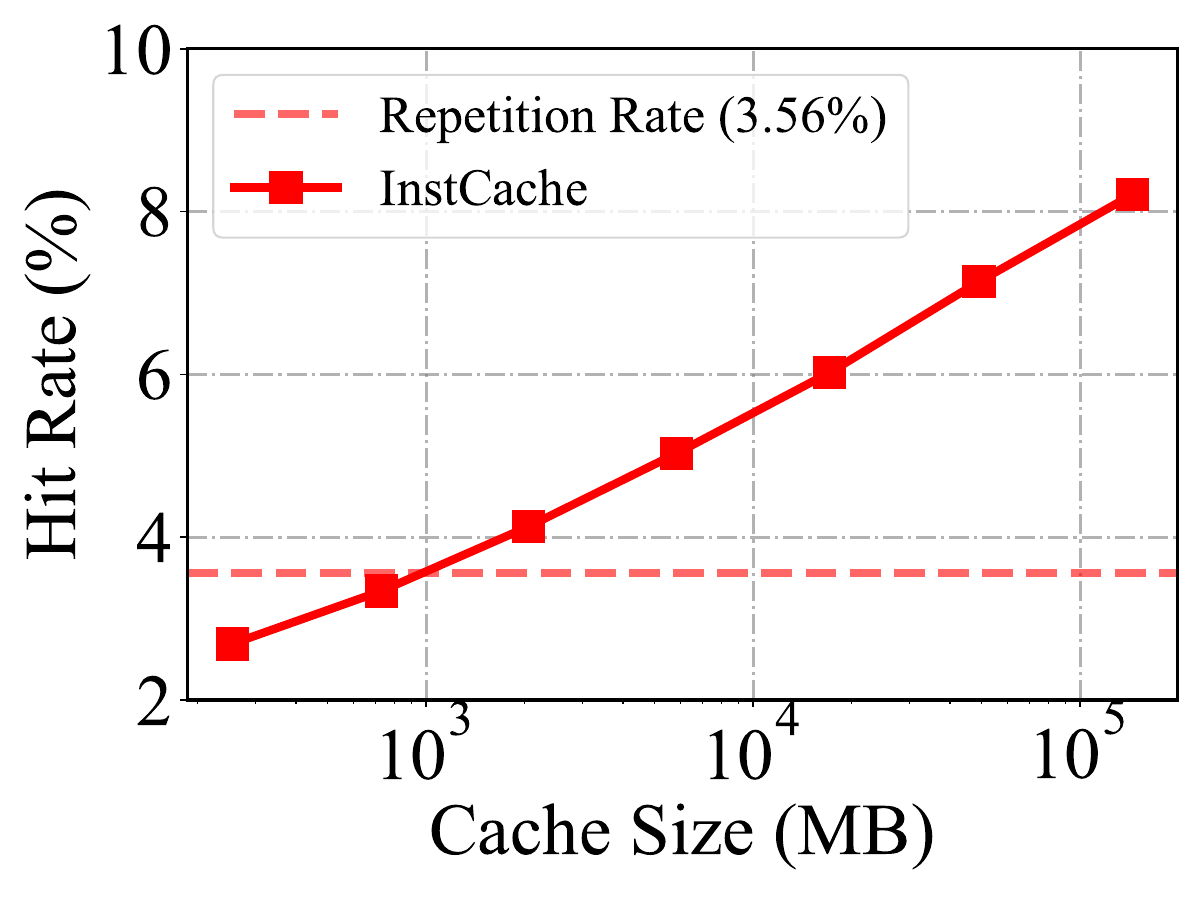}}
    \subcaptionbox{LMSys}
    {\includegraphics[width=0.231\linewidth]{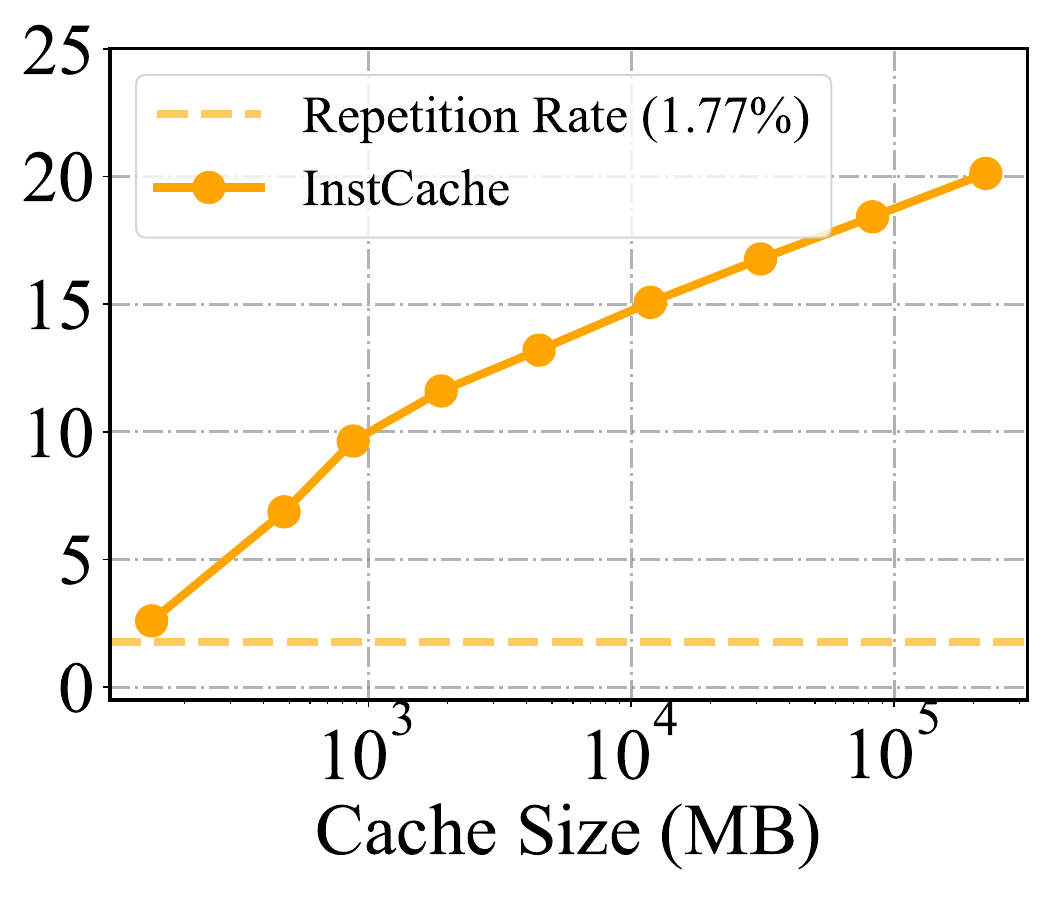}}
    \subcaptionbox{ShareGPT}
    {\includegraphics[width=0.231\linewidth]{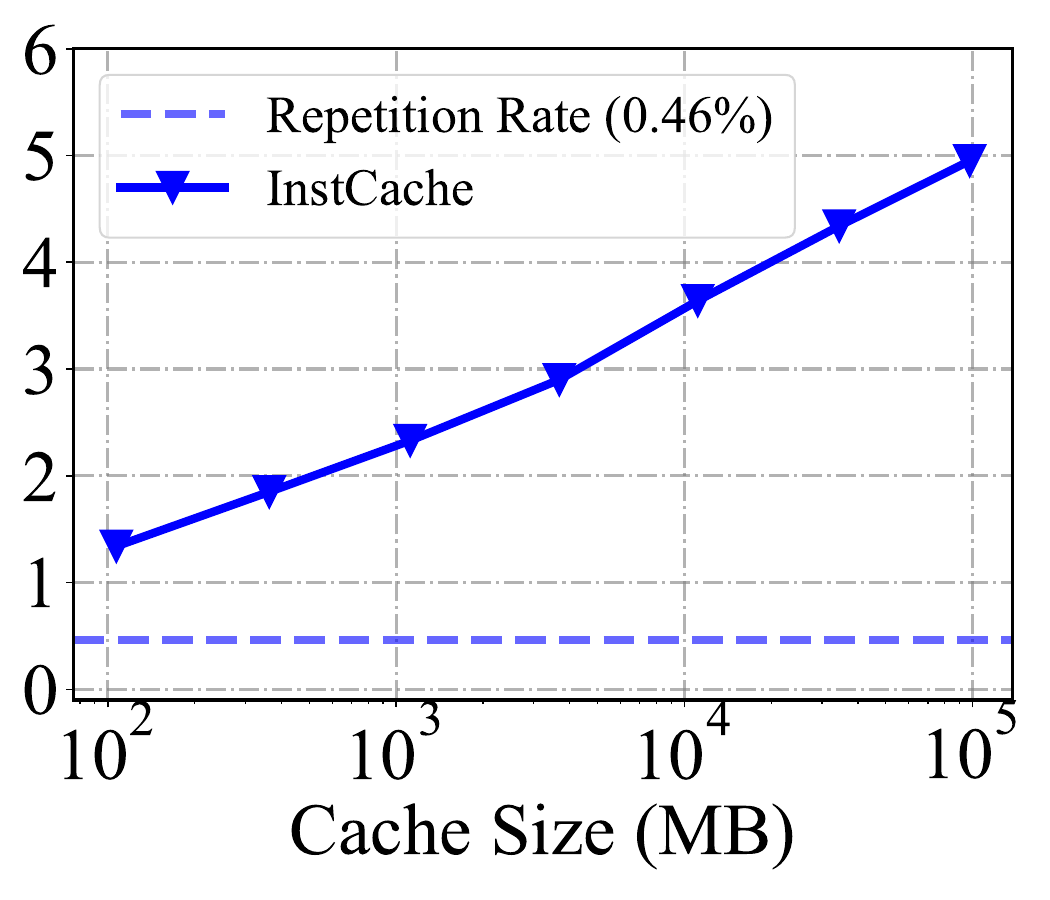}}
    \subcaptionbox{Moss}
    {\includegraphics[width=0.232\linewidth]{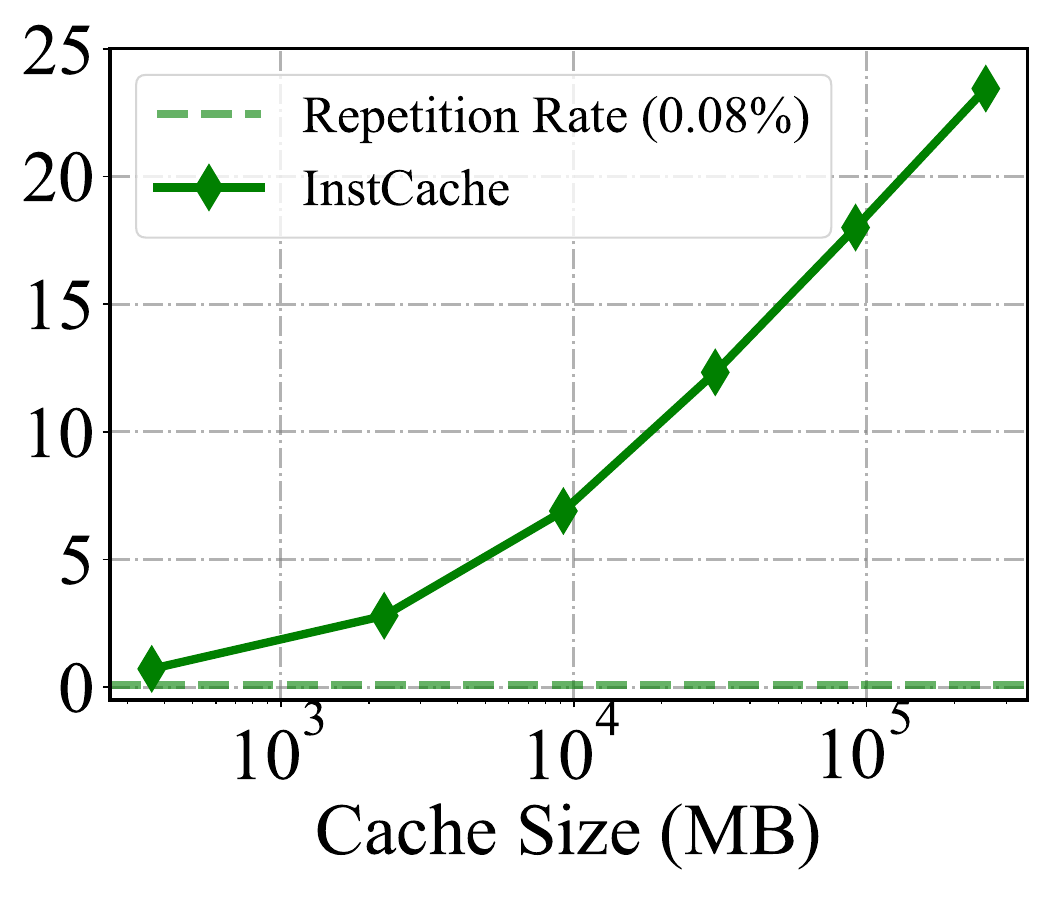}}
    \caption{Hit rates of InstCache across different datasets. The solid line represents the hit rates achieved by InstCache, while the dashed line shows the proportion of instructions in the test set that appear in the training set, indicating the upper bound of traditional caching methods.}
    \label{fig:caching_results}
\end{figure}

As illustrated in Figure \ref{fig:caching_results}, we constructed InstCache instances with NLL thresholds $\sigma$ ranging from 15 to 21 across various datasets. We then evaluate both the memory footprint and the actual hit rate for each configuration. For example, on the WildChat dataset, we achieved a hit rate of 8.2\%, which significantly surpasses the repetition rate of 3.6\% by approximately 2.3 times. Furthermore, we demonstrate hit samples of InstCache in Appendix \ref{sec:cache_details}. The hit rates of ShareGPT are comparatively lower, because ShareGPT contains only one-tenth the number of samples compared to other datasets. Additionally, the instructions collected through user sharing tend to be more challenging than those encountered in typical usage scenarios, making effective caching more difficult. 

Furthermore, the figure clearly demonstrates a continuous scaling of the hit rate with increasing cache size. This implies that for enterprise-level applications, sufficiently large storage capacities, potentially in conjunction with multi-tiered storage strategies, can unlock even greater performance gains beyond the range shown in our experiments. Finally, the high hit rates observed on the Moss dataset indicate that the LLM effectively clusters semantically similar instructions into a compact region in the text space. 

\begin{figure}[t]
    \centering
    \subcaptionbox{WildChat}
    {\includegraphics[width=0.260\linewidth]{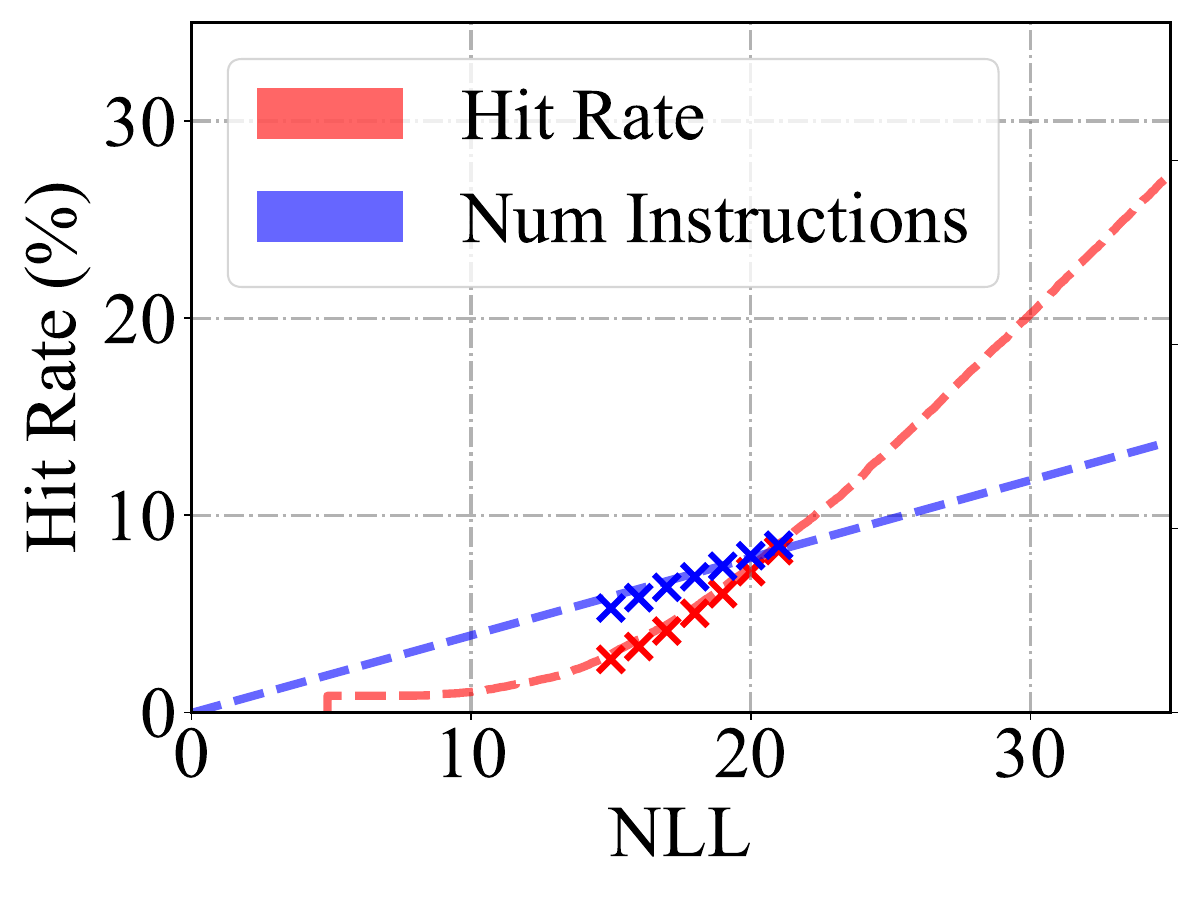}}
    \subcaptionbox{LMSys}
    {\includegraphics[width=0.228\linewidth]{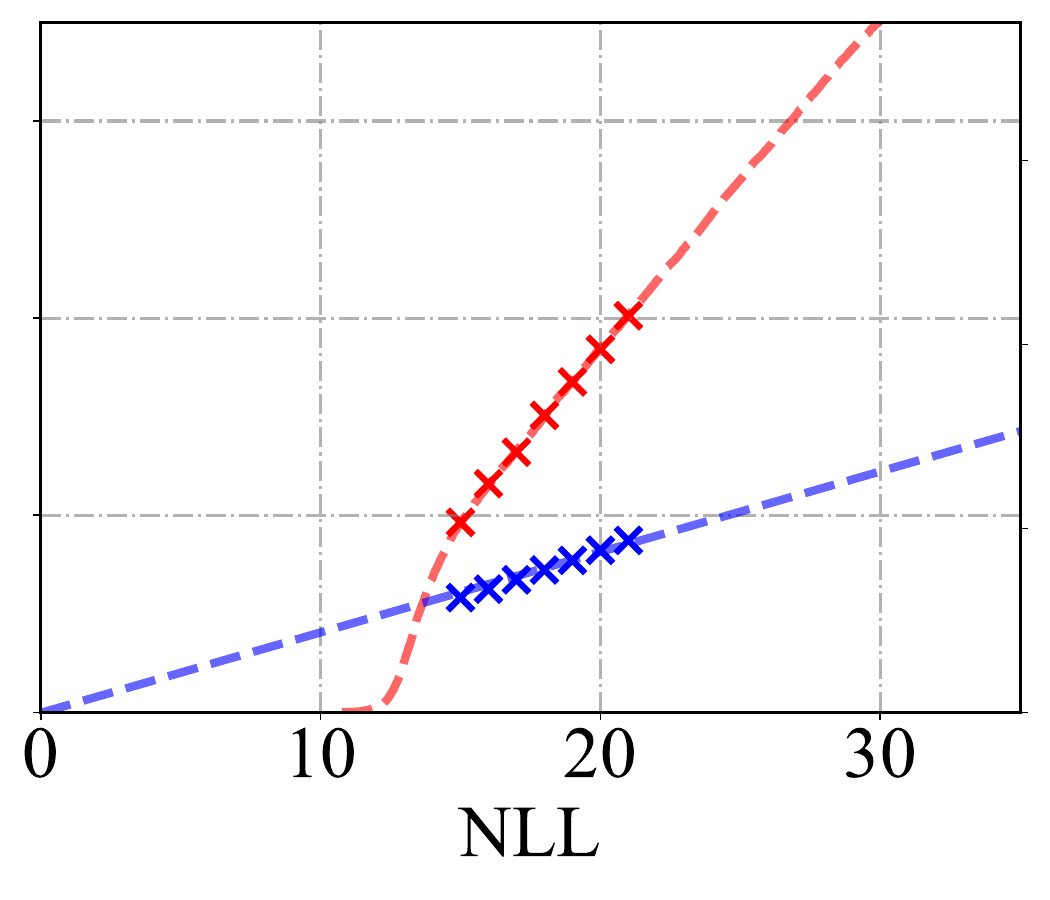}}
    \subcaptionbox{ShareGPT}
    {\includegraphics[width=0.228\linewidth]{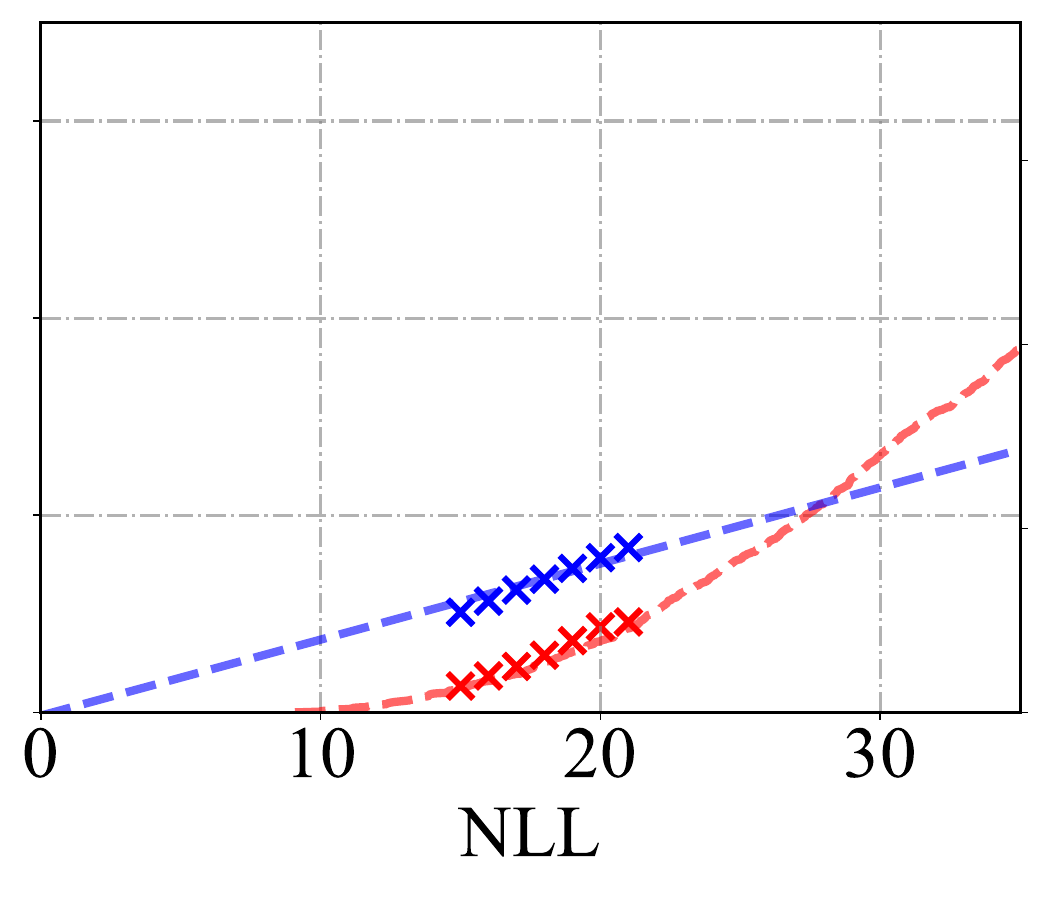}}
    \subcaptionbox{Moss}
    {\includegraphics[width=0.260\linewidth]{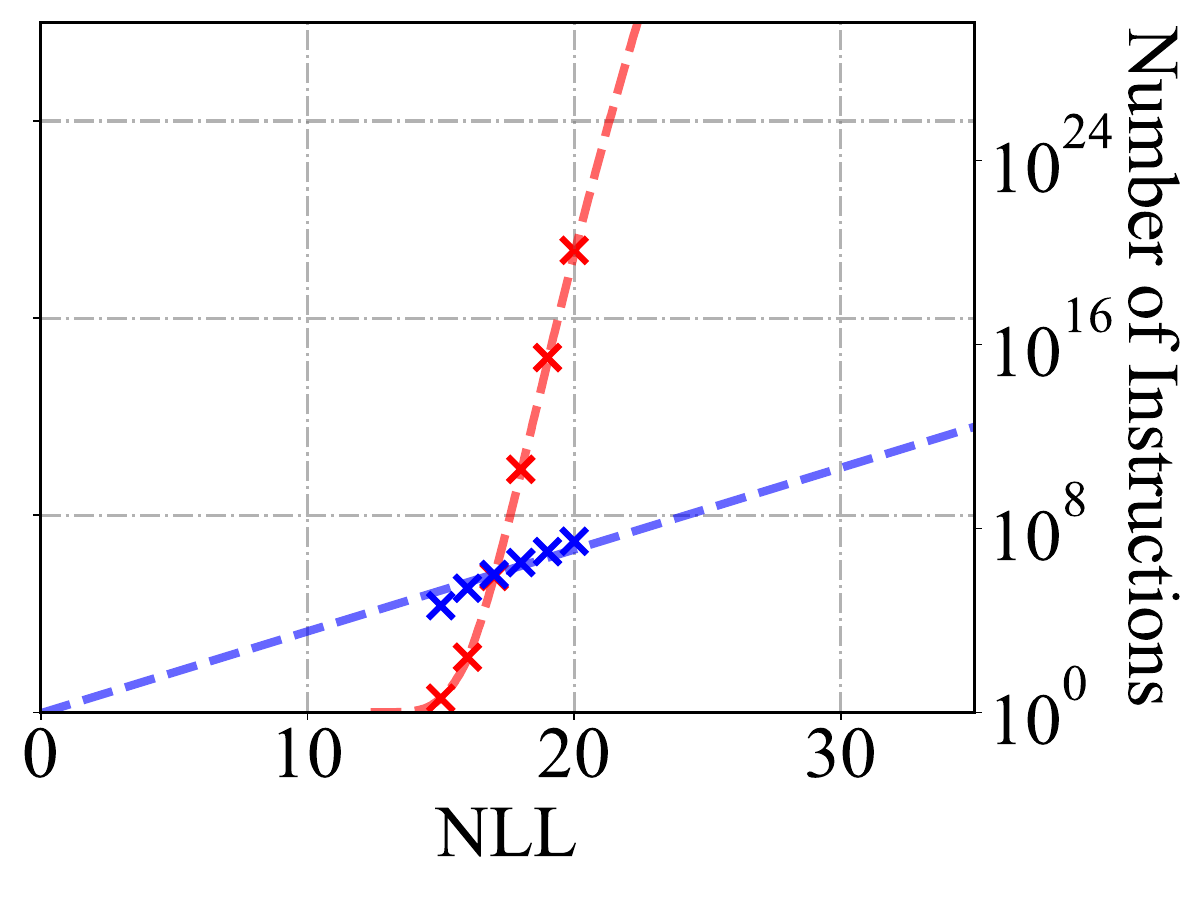}}
    \caption{Comparison between predicted and actual hit rates and instruction counts. The red and blue dashed lines represent the predicted hit rate and the number of instructions, respectively. The ”x” mark denote the actual hit rate and cache size evaluated with pre-populated InstCache.}
    \label{fig:hr_ins_nll}
\end{figure}

Moreover, it is possible for InstCache to estimate both the cache size and hit rate without the need for actual cache construction. As illustrated in the blue lines of Figure \ref{fig:hr_ins_nll}, by estimating the function parameters $\alpha, \beta$ defined in Theorem \ref{thm:text_count_estimation}, we can profile the number of instructions for InstCache accurately. Similarly, by evaluating the NLL of each instruction in the validation set using the fine-tuned LLM, we can obtain the NLL distribution and predict the hit rate according to Eq. \ref{eq:hit_rate}. As shown in the red lines of Figure\ref{fig:hr_ins_nll}, the estimated hit rate on the validation set closely aligns with the actual hit rate observed on the test set.

\subsubsection{Integration with Serving Systems}
\label{sec:integration_w_sys}

\begin{figure}[t]
    \centering
    \subcaptionbox{WildChat}
    {\includegraphics[width=0.272\linewidth]{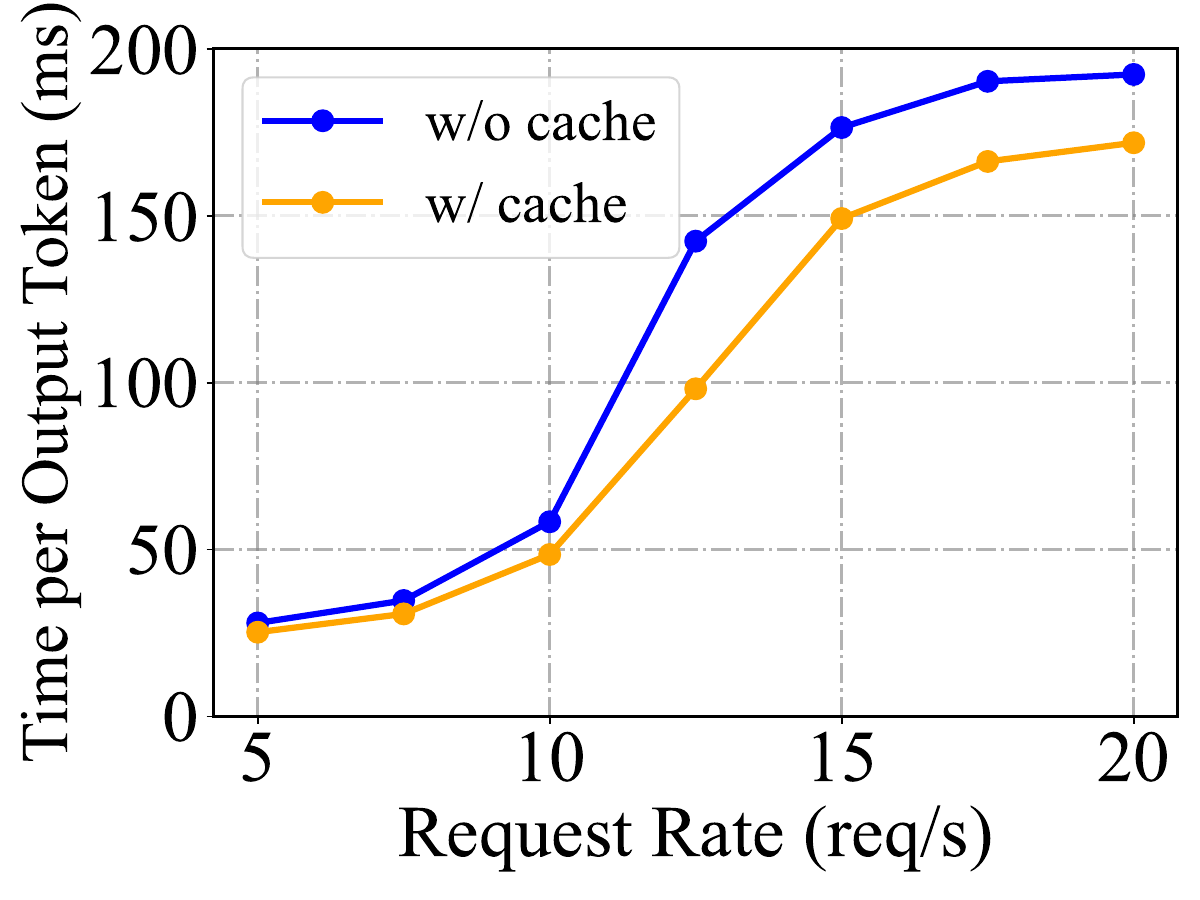}}
    \subcaptionbox{LMSys}
    {\includegraphics[width=0.231\linewidth]{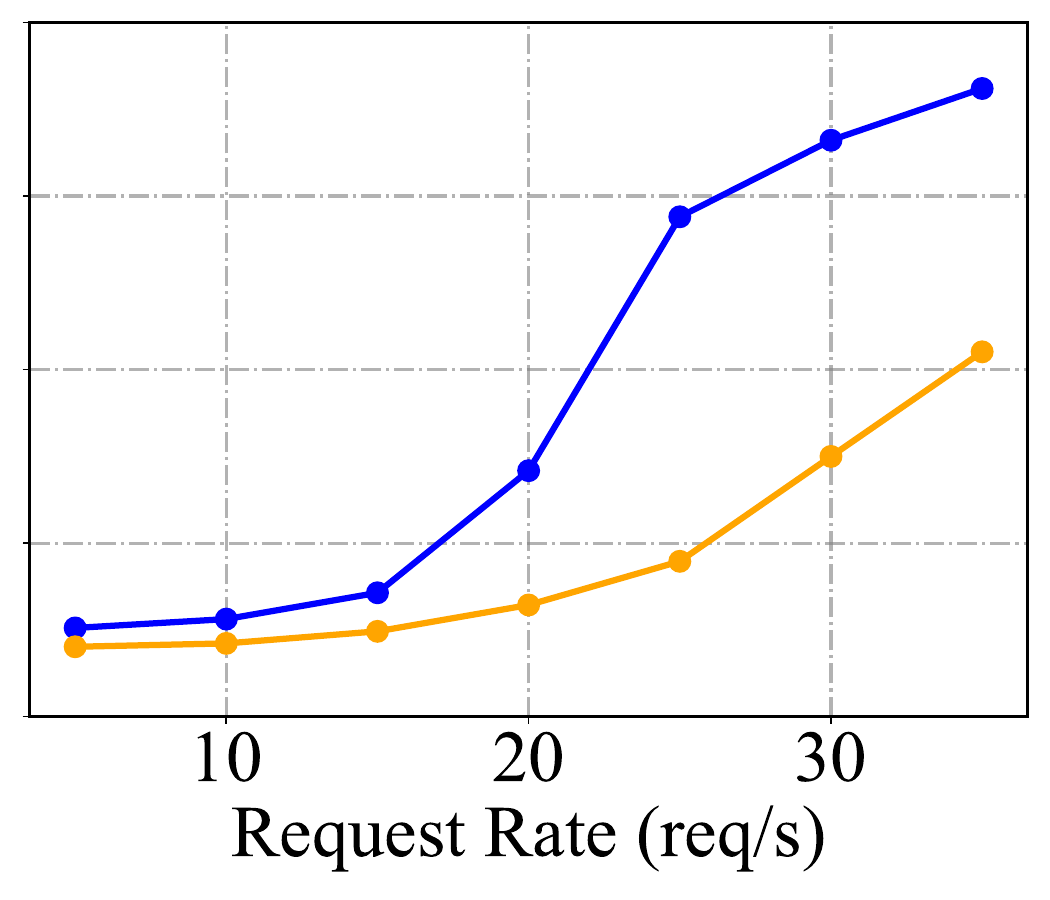}}
    \subcaptionbox{ShareGPT}
    {\includegraphics[width=0.231\linewidth]{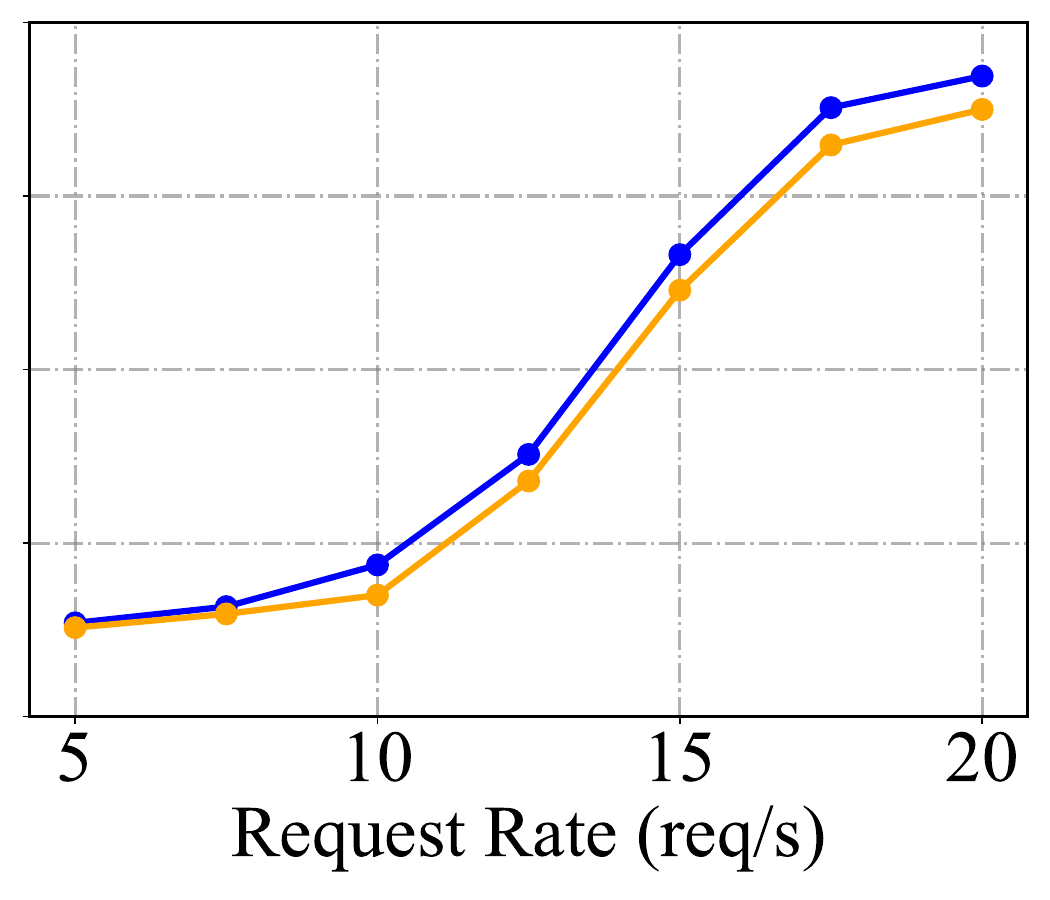}}
    \subcaptionbox{Moss}
    {\includegraphics[width=0.231\linewidth]{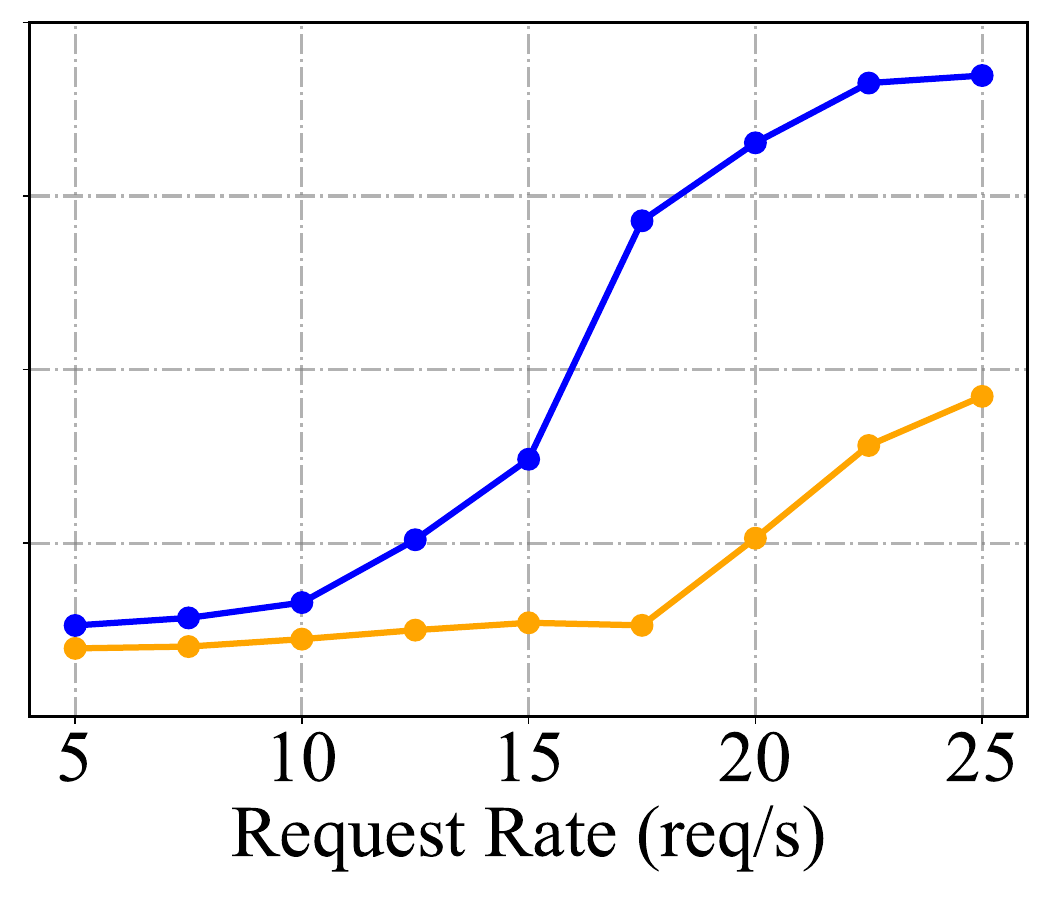}}
    \caption{Serving performance of vLLM with InstCache. We use $\sigma=21$ for InstCaches on WildChat, LMSys and ShareGPT datasets and $\sigma=20$ on the Moss dataset. Results show that InstCache becomes increasingly effective as the request rate rises.}
    \label{fig:tp}
\end{figure}

We integrate InstCache with vLLM and deploy the Qwen2.5-32B-Instruct model on two A100 GPUs for serving performance evaluation. Request arrival times are generated using a Poisson distribution with varying request rates. We evaluated the system's performance using time per output token as the serving metric. Experimental results show that InstCache improves serving performance by up to 10.4\%, 42.0\%, 4.9\%, and 50.0\% on the WildChat, LMSys, ShareGPT, and Moss datasets, respectively. The relatively modest improvement on ShareGPT is caused by its smaller dataset size, which limits the cache hit rate of InstCache. Across all datasets, performance gains increase with higher request rates, as cache hits effectively reduce the system load proportionally. Consequently, InstCache delivers greater performance improvements under heavier traffic.

\subsubsection{Cost of Pre-population}
\label{sec:prepopulation_cost}

\begin{figure}[htbp]
  \begin{minipage}[t]{0.64\textwidth}
    \centering
    \subcaptionbox{Pre-population Time\label{fig:prepopulation_time}}
    {\includegraphics[width=0.49\linewidth]{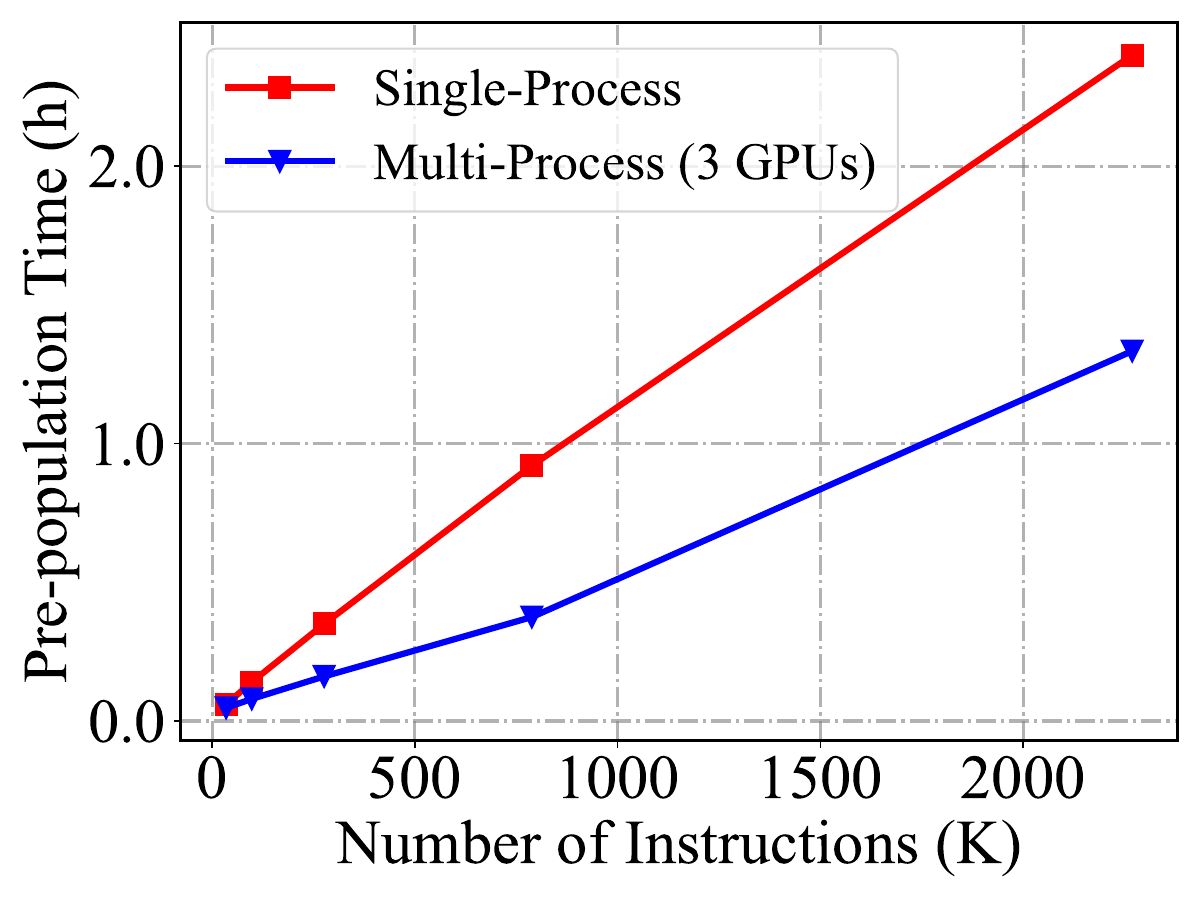}}
    % \hspace{0.4cm}
    \hfill
    \subcaptionbox{Maximum KVCache Memory\label{fig:max_kvcache_mem}}
    {\includegraphics[width=0.49\linewidth]{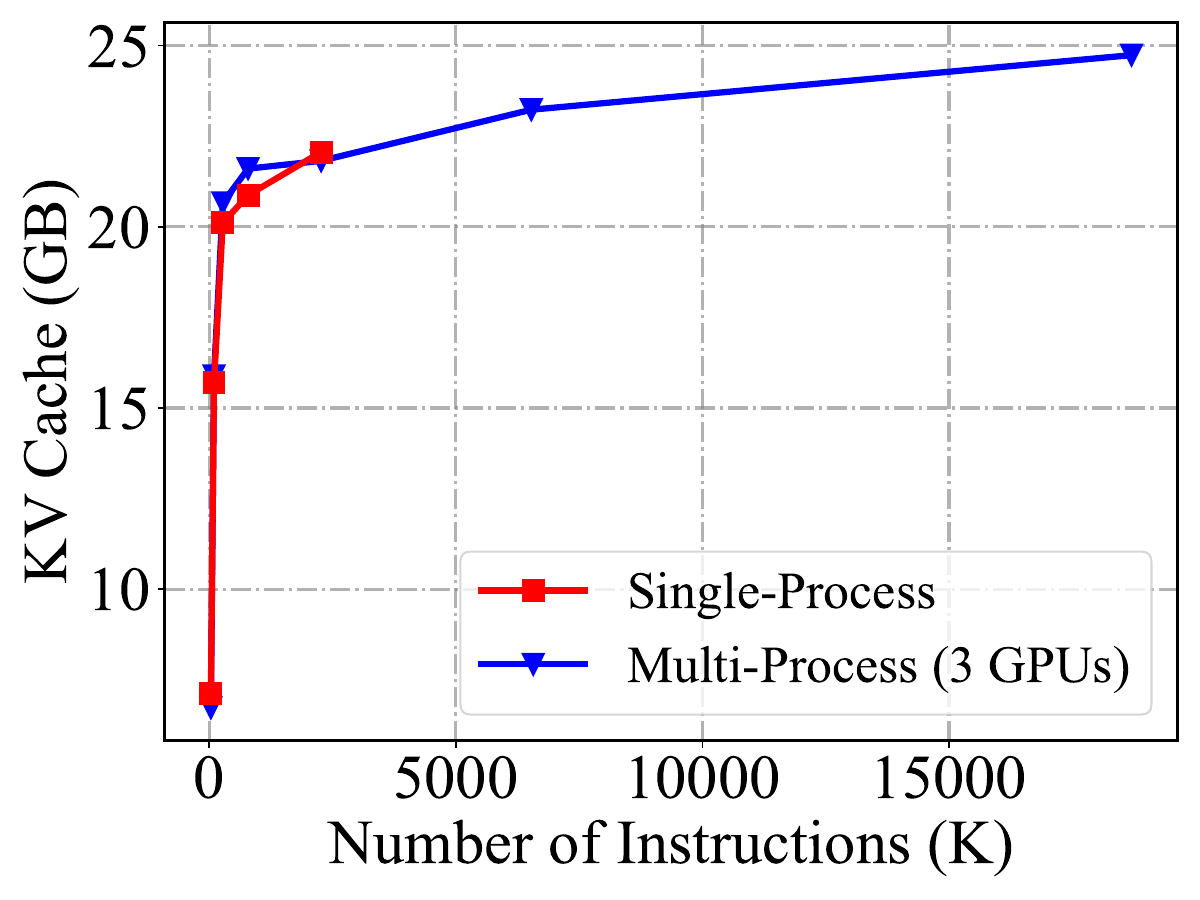}}
    \caption{Cost of Pre-population on the WildChat dataset. Subfigure (a) illustrates the pre-population time across varying cache sizes. Subfigure (b) depicts the maximum KVCache memory usage during pre-population for different cache sizes.}
    \label{fig:cost_prepopulation}
  \end{minipage}
  \hfill
  \centering
  \begin{minipage}[t]{0.32\textwidth}
    \centering
    \subcaptionbox{Effect of Distribution Shift}
    {\includegraphics[width=\linewidth]{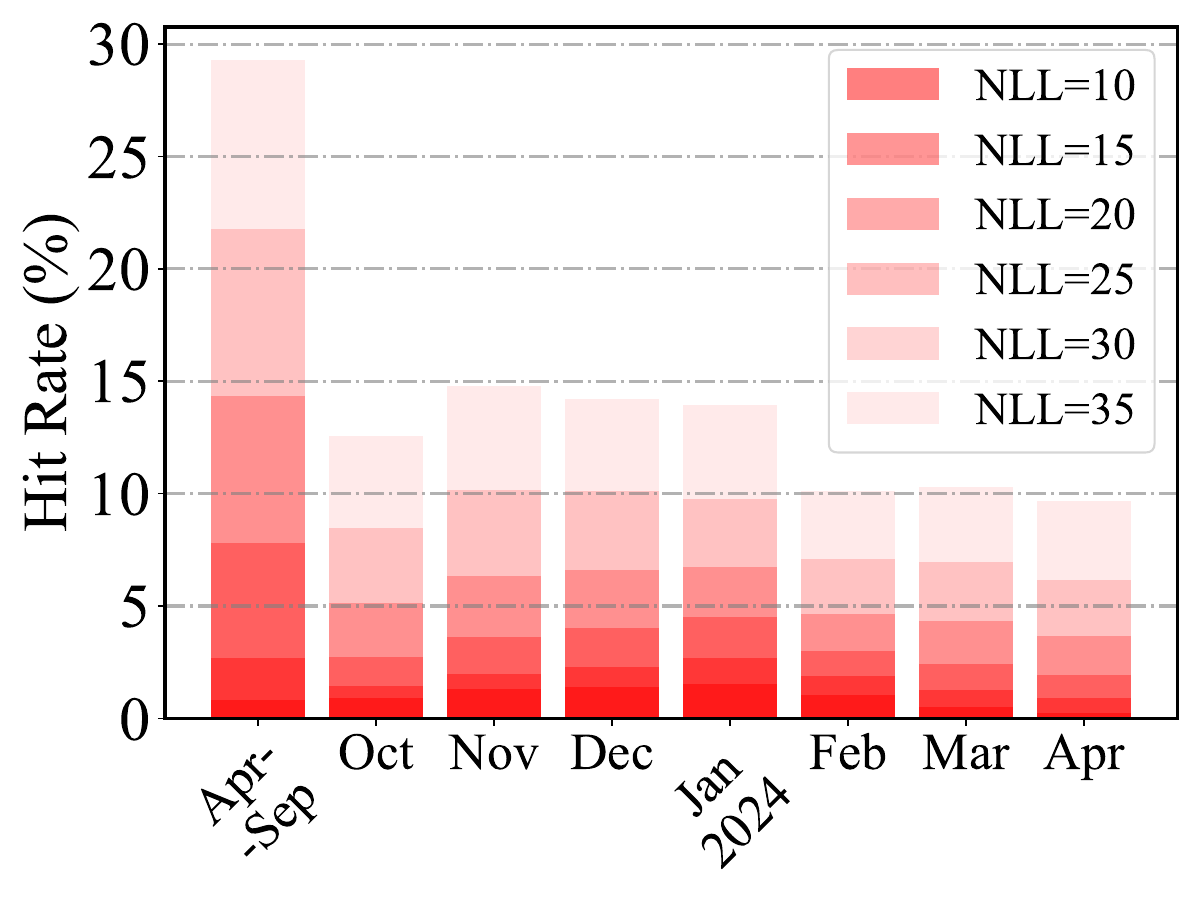}}
    \caption{Evaluation of the impact of instruction distribution shift on hit rate.}
    \label{fig:dynamic}
  \end{minipage}
\end{figure}

Figure \ref{fig:prepopulation_time} shows the pre-population time with varying numbers of instruction on WildChat dataset. As observed, the computational cost scales linearly with the number of instructions. Notably, the distributed approach has a fixed overhead incurred when preparing a small V-ary tree, which must be constructed on a single GPU. Additionally, task distribution across GPUs also introduces a modest load-balancing overhead. Therefore, as the task size grows, these overheads are effectively amortized, and the efficiency of distributed pre-population gradually approaches the theoretical speedup corresponding to the number of GPUs.

Figure \ref{fig:max_kvcache_mem} demonstrates the relationship between the GPU memory footprint of the KVCache in our pre-population system and the number of instructions. As the number of instructions increases, the growth rate of the KVCache memory usage on the GPU gradually slow down, validating the effectiveness of our deep first searching strategy, which minimizes the overhead associated with preserving KVCache for existing nodes in the V-ary tree. Moreover, the KVCache memory overhead can be further reduced by decreasing the batch size.

\subsubsection{Effect of Distribution Shift}
\label{sec:distribution_shift}

As user instruction distributions may shift over time, the hit rate of a static InstCache will gradually decline. However, benefited from the semantic capability of LLM for cache construction, InstCache can still capture instruction similarity despite temporal drift. Since the WildChat dataset includes timestamps spanning an entire year, we first train the LLM on the data from the first six months. After that, we evaluate the hit rate with different NLLs on the test datasets from the first six months and the subsequent months. As illustrated in Figure \ref{fig:dynamic}, the experimental results indicate that as the temporal gap between the training and testing datasets increases, InstCache performance gradually degrades. Nevertheless, the decline remains steady over time, proving that most popular instructions tend to remain stable and do not change frequently.

\section{Discussion}

Since InstCache is a static cache, gradual shifts in the instruction distribution can cause a decline in hit rates, as demonstrated in Section \ref{sec:distribution_shift}. Therefore, periodic cache updates are necessary to maintain an up-to-date representation of the instruction distribution. Thanks to the highly optimized tree-searching algorithm, the pre-population overhead remains relatively low. Moreover, existing responses to observed instructions and previously pre-populated instructions can be reused in subsequent pre-population cycles if they share common instructions, further reducing the overall cost.

In addition to common conversational applications, InstCache also supports tasks that involve post-processing of user prompts, such as Retrieval-Augmented Generation (RAG)\citep{RAG}. For RAG tasks, InstCache only predicts user-written instructions and then appends additional documents retrieved from the database to generate responses. For use cases that requires sampling different responses to the same instruction, InstCache can be used at the initial stage, while subsequent instructions from the same user within a short time period can bypass the cache. Of course, instruction-response caching mechanisms are not proper for instructions with high timeliness.

Finally, extra preprocessing can be necessary before fitting the instruction distribution with observed instructions. In this paper, we remove any instructions containing personal information, such as names, locations, and other sensitive details. In real-world deployment scenarios, harmful or personal instructions can be filtered out after pre-population so as to minimize the risk of data leakage.

\section{Conclusion}

In this paper, we address the challenge of dissipated locality in instruction-response caching systems by leveraging the semantic capability of LLM to reorder the text space and develop spatial locality. Utilizing the spatial locality, we can predict potential instructions and construct the InstCache. Furthermore, we demonstrate that the hit rate and cache size of InstCache can be predicted. To efficiently pre-populate InstCache, we develop a depth-first V-ary tree search algorithm. Experimental results validate the effectiveness of InstCache. In the future, we will develop a distributed implementation of the caching system on larger computer clusters to further enhance the performance.

\bibliographystyle{plain}
\bibliography{paper}

\begin{thebibliography}{10}

\bibitem{alignment_tax}
Amanda Askell, Yuntao Bai, Anna Chen, Dawn Drain, Deep Ganguli, Tom Henighan, Andy Jones, Nicholas Joseph, Benjamin Mann, Nova DasSarma, Nelson Elhage, Zac Hatfield{-}Dodds, Danny Hernandez, Jackson Kernion, Kamal Ndousse, Catherine Olsson, Dario Amodei, Tom~B. Brown, Jack Clark, Sam McCandlish, Chris Olah, and Jared Kaplan.
\newblock A general language assistant as a laboratory for alignment.
\newblock {\em CoRR}, abs/2112.00861, 2021.

\bibitem{spire2003}
Ricardo~A. Baeza{-}Yates and Felipe Saint{-}Jean.
\newblock A three level search engine index based in query log distribution.
\newblock In Mario~A. Nascimento, Edleno~Silva de~Moura, and Arlindo~L. Oliveira, editors, {\em String Processing and Information Retrieval, 10th International Symposium, {SPIRE} 2003, Manaus, Brazil, October 8-10, 2003, Proceedings}, volume 2857 of {\em Lecture Notes in Computer Science}, pages 56--65. Springer, 2003.

\bibitem{GPTCache}
Fu~Bang.
\newblock {GPTC}ache: An open-source semantic cache for {LLM} applications enabling faster answers and cost savings.
\newblock In Liling Tan, Dmitrijs Milajevs, Geeticka Chauhan, Jeremy Gwinnup, and Elijah Rippeth, editors, {\em Proceedings of the 3rd Workshop for Natural Language Processing Open Source Software (NLP-OSS 2023)}, pages 212--218, Singapore, December 2023. Association for Computational Linguistics.

\bibitem{arena}
Wei{-}Lin Chiang, Lianmin Zheng, Ying Sheng, Anastasios~Nikolas Angelopoulos, Tianle Li, Dacheng Li, Banghua Zhu, Hao Zhang, Michael~I. Jordan, Joseph~E. Gonzalez, and Ion Stoica.
\newblock Chatbot arena: An open platform for evaluating llms by human preference.
\newblock In {\em Forty-first International Conference on Machine Learning, {ICML} 2024, Vienna, Austria, July 21-27, 2024}. OpenReview.net, 2024.

\bibitem{mla}
DeepSeek{-}AI, Aixin Liu, Bei Feng, Bin Wang, Bingxuan Wang, Bo~Liu, Chenggang Zhao, Chengqi Deng, Chong Ruan, Damai Dai, Daya Guo, Dejian Yang, Deli Chen, Dongjie Ji, Erhang Li, Fangyun Lin, Fuli Luo, Guangbo Hao, Guanting Chen, Guowei Li, Hao Zhang, Hanwei Xu, Hao Yang, Haowei Zhang, Honghui Ding, Huajian Xin, Huazuo Gao, Hui Li, Hui Qu, J.~L. Cai, Jian Liang, Jianzhong Guo, Jiaqi Ni, Jiashi Li, Jin Chen, Jingyang Yuan, Junjie Qiu, Junxiao Song, Kai Dong, Kaige Gao, Kang Guan, Lean Wang, Lecong Zhang, Lei Xu, Leyi Xia, Liang Zhao, Liyue Zhang, Meng Li, Miaojun Wang, Mingchuan Zhang, Minghua Zhang, Minghui Tang, Mingming Li, Ning Tian, Panpan Huang, Peiyi Wang, Peng Zhang, Qihao Zhu, Qinyu Chen, Qiushi Du, R.~J. Chen, R.~L. Jin, Ruiqi Ge, Ruizhe Pan, Runxin Xu, Ruyi Chen, S.~S. Li, Shanghao Lu, Shangyan Zhou, Shanhuang Chen, Shaoqing Wu, Shengfeng Ye, Shirong Ma, Shiyu Wang, Shuang Zhou, Shuiping Yu, Shunfeng Zhou, Size Zheng, Tao Wang, Tian Pei, Tian Yuan, Tianyu Sun, W.~L. Xiao, Wangding Zeng, Wei An, Wen
  Liu, Wenfeng Liang, Wenjun Gao, Wentao Zhang, X.~Q. Li, Xiangyue Jin, Xianzu Wang, Xiao Bi, Xiaodong Liu, Xiaohan Wang, Xiaojin Shen, Xiaokang Chen, Xiaosha Chen, Xiaotao Nie, and Xiaowen Sun.
\newblock Deepseek-v2: {A} strong, economical, and efficient mixture-of-experts language model.
\newblock {\em CoRR}, abs/2405.04434, 2024.

\bibitem{deepseek_v3}
DeepSeek{-}AI, Aixin Liu, Bei Feng, Bing Xue, Bingxuan Wang, Bochao Wu, Chengda Lu, Chenggang Zhao, Chengqi Deng, Chenyu Zhang, Chong Ruan, Damai Dai, Daya Guo, Dejian Yang, Deli Chen, Dongjie Ji, Erhang Li, Fangyun Lin, Fucong Dai, Fuli Luo, Guangbo Hao, Guanting Chen, Guowei Li, H.~Zhang, Han Bao, Hanwei Xu, Haocheng Wang, Haowei Zhang, Honghui Ding, Huajian Xin, Huazuo Gao, Hui Li, Hui Qu, J.~L. Cai, Jian Liang, Jianzhong Guo, Jiaqi Ni, Jiashi Li, Jiawei Wang, Jin Chen, Jingchang Chen, Jingyang Yuan, Junjie Qiu, Junlong Li, Junxiao Song, Kai Dong, Kai Hu, Kaige Gao, Kang Guan, Kexin Huang, Kuai Yu, Lean Wang, Lecong Zhang, Lei Xu, Leyi Xia, Liang Zhao, Litong Wang, Liyue Zhang, Meng Li, Miaojun Wang, Mingchuan Zhang, Minghua Zhang, Minghui Tang, Mingming Li, Ning Tian, Panpan Huang, Peiyi Wang, Peng Zhang, Qiancheng Wang, Qihao Zhu, Qinyu Chen, Qiushi Du, R.~J. Chen, R.~L. Jin, Ruiqi Ge, Ruisong Zhang, Ruizhe Pan, Runji Wang, Runxin Xu, Ruoyu Zhang, Ruyi Chen, S.~S. Li, Shanghao Lu, Shangyan Zhou,
  Shanhuang Chen, Shaoqing Wu, Shengfeng Ye, Shengfeng Ye, Shirong Ma, Shiyu Wang, Shuang Zhou, Shuiping Yu, Shunfeng Zhou, Shuting Pan, T.~Wang, Tao Yun, Tian Pei, Tianyu Sun, W.~L. Xiao, and Wangding Zeng.
\newblock Deepseek-v3 technical report.
\newblock {\em CoRR}, abs/2412.19437, 2024.

\bibitem{llama3}
Abhimanyu Dubey, Abhinav Jauhri, Abhinav Pandey, Abhishek Kadian, Ahmad Al{-}Dahle, Aiesha Letman, Akhil Mathur, Alan Schelten, Amy Yang, Angela Fan, Anirudh Goyal, Anthony Hartshorn, Aobo Yang, Archi Mitra, Archie Sravankumar, Artem Korenev, Arthur Hinsvark, Arun Rao, Aston Zhang, Aur{\'{e}}lien Rodriguez, Austen Gregerson, Ava Spataru, Baptiste Rozi{\`{e}}re, Bethany Biron, Binh Tang, Bobbie Chern, Charlotte Caucheteux, Chaya Nayak, Chloe Bi, Chris Marra, Chris McConnell, Christian Keller, Christophe Touret, Chunyang Wu, Corinne Wong, Cristian~Canton Ferrer, Cyrus Nikolaidis, Damien Allonsius, Daniel Song, Danielle Pintz, Danny Livshits, David Esiobu, Dhruv Choudhary, Dhruv Mahajan, Diego Garcia{-}Olano, Diego Perino, Dieuwke Hupkes, Egor Lakomkin, Ehab AlBadawy, Elina Lobanova, Emily Dinan, Eric~Michael Smith, Filip Radenovic, Frank Zhang, Gabriel Synnaeve, Gabrielle Lee, Georgia~Lewis Anderson, Graeme Nail, Gr{\'{e}}goire Mialon, Guan Pang, Guillem Cucurell, Hailey Nguyen, Hannah Korevaar, Hu~Xu, Hugo
  Touvron, Iliyan Zarov, Imanol~Arrieta Ibarra, Isabel~M. Kloumann, Ishan Misra, Ivan Evtimov, Jade Copet, Jaewon Lee, Jan Geffert, Jana Vranes, Jason Park, Jay Mahadeokar, Jeet Shah, Jelmer van~der Linde, Jennifer Billock, Jenny Hong, Jenya Lee, Jeremy Fu, Jianfeng Chi, Jianyu Huang, Jiawen Liu, Jie Wang, Jiecao Yu, Joanna Bitton, Joe Spisak, Jongsoo Park, Joseph Rocca, Joshua Johnstun, Joshua Saxe, Junteng Jia, Kalyan~Vasuden Alwala, Kartikeya Upasani, Kate Plawiak, Ke~Li, Kenneth Heafield, Kevin Stone, and et~al.
\newblock The llama 3 herd of models.
\newblock {\em CoRR}, abs/2407.21783, 2024.

\bibitem{tois2006}
Tiziano Fagni, Raffaele Perego, Fabrizio Silvestri, and Salvatore Orlando.
\newblock Boosting the performance of web search engines: Caching and prefetching query results by exploiting historical usage data.
\newblock {\em {ACM} Trans. Inf. Syst.}, 24(1):51--78, 2006.

\bibitem{attentionstore}
Bin Gao, Zhuomin He, Puru Sharma, Qingxuan Kang, Djordje Jevdjic, Junbo Deng, Xingkun Yang, Zhou Yu, and Pengfei Zuo.
\newblock Attentionstore: Cost-effective attention reuse across multi-turn conversations in large language model serving.
\newblock {\em CoRR}, abs/2403.19708, 2024.

\bibitem{RAG}
Yunfan Gao, Yun Xiong, Xinyu Gao, Kangxiang Jia, Jinliu Pan, Yuxi Bi, Yi~Dai, Jiawei Sun, Qianyu Guo, Meng Wang, and Haofen Wang.
\newblock Retrieval-augmented generation for large language models: {A} survey.
\newblock {\em CoRR}, abs/2312.10997, 2023.

\bibitem{meancache}
Waris Gill, Mohamed Elidrisi, Pallavi Kalapatapu, Ali Anwar, and Muhammad~Ali Gulzar.
\newblock Privacy-aware semantic cache for large language models.
\newblock {\em CoRR}, abs/2403.19708, 2024.

\bibitem{vllm}
Woosuk Kwon, Zhuohan Li, Siyuan Zhuang, Ying Sheng, Lianmin Zheng, Cody~Hao Yu, Joseph Gonzalez, Hao Zhang, and Ion Stoica.
\newblock Efficient memory management for large language model serving with pagedattention.
\newblock In Jason Flinn, Margo~I. Seltzer, Peter Druschel, Antoine Kaufmann, and Jonathan Mace, editors, {\em Proceedings of the 29th Symposium on Operating Systems Principles, {SOSP} 2023, Koblenz, Germany, October 23-26, 2023}, pages 611--626. {ACM}, 2023.

\bibitem{www2003}
Ronny Lempel and Shlomo Moran.
\newblock Predictive caching and prefetching of query results in search engines.
\newblock In Guszt{\'{a}}v Hencsey, Bebo White, Yih{-}Farn~Robin Chen, L{\'{a}}szl{\'{o}} Kov{\'{a}}cs, and Steve Lawrence, editors, {\em Proceedings of the Twelfth International World Wide Web Conference, {WWW} 2003, Budapest, Hungary, May 20-24, 2003}, pages 19--28. {ACM}, 2003.

\bibitem{scalm}
Jiaxing Li, Chi Xu, Feng Wang, Isaac~M. von Riedemann, Cong Zhang, and Jiangchuan Liu.
\newblock {SCALM:} towards semantic caching for automated chat services with large language models.
\newblock In {\em 32nd {IEEE/ACM} International Symposium on Quality of Service, IWQoS 2024, Guangzhou, China, June 19-21, 2024}, pages 1--10. {IEEE}, 2024.

\bibitem{www2005}
Xiaohui Long and Torsten Suel.
\newblock Three-level caching for efficient query processing in large web search engines.
\newblock In Allan Ellis and Tatsuya Hagino, editors, {\em Proceedings of the 14th international conference on World Wide Web, {WWW} 2005, Chiba, Japan, May 10-14, 2005}, pages 257--266. {ACM}, 2005.

\bibitem{hpdc2010}
Mauricio Mar{\'{\i}}n, Veronica Gil{-}Costa, and Carlos G{\'{o}}mez{-}Pantoja.
\newblock New caching techniques for web search engines.
\newblock In Salim Hariri and Kate Keahey, editors, {\em Proceedings of the 19th {ACM} International Symposium on High Performance Distributed Computing, {HPDC} 2010, Chicago, Illinois, USA, June 21-25, 2010}, pages 215--226. {ACM}, 2010.

\bibitem{querycache}
Evangelos~P. Markatos.
\newblock On caching search engine query results.
\newblock {\em Comput. Commun.}, 24(2):137--143, 2001.

\bibitem{contextcache}
Ramaswami Mohandoss.
\newblock Context-based semantic caching for llm applications.
\newblock In {\em 2024 IEEE Conference on Artificial Intelligence (CAI)}, pages 371--376. IEEE, 2024.

\bibitem{openai}
OpenAI.
\newblock Openai chatgpt, 2022.

\bibitem{input_gen}
Zafaryab Rasool, Scott Barnett, David Willie, Stefanus Kurniawan, Sherwin Balugo, Srikanth Thudumu, and Mohamed~Almorsy Abdelrazek.
\newblock Llms for test input generation for semantic caches.
\newblock {\em CoRR}, abs/2401.08138, 2024.

\bibitem{sbert}
Nils Reimers and Iryna Gurevych.
\newblock Sentence-bert: Sentence embeddings using siamese bert-networks.
\newblock In Kentaro Inui, Jing Jiang, Vincent Ng, and Xiaojun Wan, editors, {\em Proceedings of the 2019 Conference on Empirical Methods in Natural Language Processing and the 9th International Joint Conference on Natural Language Processing, {EMNLP-IJCNLP} 2019, Hong Kong, China, November 3-7, 2019}, pages 3980--3990. Association for Computational Linguistics, 2019.

\bibitem{sigir2001}
Patricia~Correia Saraiva, Edleno~Silva de~Moura, Rodrigo~C. Fonseca, Wagner~Meira Jr., Berthier~A. Ribeiro{-}Neto, and Nivio Ziviani.
\newblock Rank-preserving two-level caching for scalable search engines.
\newblock In W.~Bruce Croft, David~J. Harper, Donald~H. Kraft, and Justin Zobel, editors, {\em {SIGIR} 2001: Proceedings of the 24th Annual International {ACM} {SIGIR} Conference on Research and Development in Information Retrieval, September 9-13, 2001, New Orleans, Louisiana, {USA}}, pages 51--58. {ACM}, 2001.

\bibitem{dataleakage}
Linke Song, Zixuan Pang, Wenhao Wang, Zihao Wang, XiaoFeng Wang, Hongbo Chen, Wei Song, Yier Jin, Dan Meng, and Rui Hou.
\newblock The early bird catches the leak: Unveiling timing side channels in {LLM} serving systems.
\newblock {\em CoRR}, abs/2409.20002, 2024.

\bibitem{moss}
Tianxiang Sun, Xiaotian Zhang, Zhengfu He, Peng Li, Qinyuan Cheng, Xiangyang Liu, Hang Yan, Yunfan Shao, Qiong Tang, Shiduo Zhang, Xingjian Zhao, Ke~Chen, Yining Zheng, Zhejian Zhou, Ruixiao Li, Jun Zhan, Yunhua Zhou, Linyang Li, Xiaogui Yang, Lingling Wu, Zhangyue Yin, Xuanjing Huang, Yu{-}Gang Jiang, and Xipeng Qiu.
\newblock {MOSS:} an open conversational large language model.
\newblock {\em Mach. Intell. Res.}, 21(5):888--905, 2024.

\bibitem{sharegpt}
Team.
\newblock Sharegpt, 2023.

\bibitem{attentionsink}
Guangxuan Xiao, Yuandong Tian, Beidi Chen, Song Han, and Mike Lewis.
\newblock Efficient streaming language models with attention sinks.
\newblock In {\em The Twelfth International Conference on Learning Representations, {ICLR} 2024, Vienna, Austria, May 7-11, 2024}. OpenReview.net, 2024.

\bibitem{querylocality}
Yinglian Xie and David~R. O'Hallaron.
\newblock Locality in search engine queries and its implications for caching.
\newblock In {\em Proceedings of {IEEE} {INFOCOM}}, pages 1238--1247. {IEEE} Computer Society, 2002.

\bibitem{qwen2.5}
An~Yang, Baosong Yang, Beichen Zhang, Binyuan Hui, Bo~Zheng, Bowen Yu, Chengyuan Li, Dayiheng Liu, Fei Huang, Haoran Wei, Huan Lin, Jian Yang, Jianhong Tu, Jianwei Zhang, Jianxin Yang, Jiaxi Yang, Jingren Zhou, Junyang Lin, Kai Dang, Keming Lu, Keqin Bao, Kexin Yang, Le~Yu, Mei Li, Mingfeng Xue, Pei Zhang, Qin Zhu, Rui Men, Runji Lin, Tianhao Li, Tingyu Xia, Xingzhang Ren, Xuancheng Ren, Yang Fan, Yang Su, Yichang Zhang, Yu~Wan, Yuqiong Liu, Zeyu Cui, Zhenru Zhang, and Zihan Qiu.
\newblock Qwen2.5 technical report.
\newblock {\em CoRR}, abs/2412.15115, 2024.

\bibitem{chunkattention}
Lu~Ye, Ze~Tao, Yong Huang, and Yang Li.
\newblock Chunkattention: Efficient self-attention with prefix-aware {KV} cache and two-phase partition.
\newblock In Lun{-}Wei Ku, Andre Martins, and Vivek Srikumar, editors, {\em Proceedings of the 62nd Annual Meeting of the Association for Computational Linguistics (Volume 1: Long Papers), {ACL} 2024, Bangkok, Thailand, August 11-16, 2024}, pages 11608--11620. Association for Computational Linguistics, 2024.

\bibitem{www2008}
Jiangong Zhang, Xiaohui Long, and Torsten Suel.
\newblock Performance of compressed inverted list caching in search engines.
\newblock In Jinpeng Huai, Robin Chen, Hsiao{-}Wuen Hon, Yunhao Liu, Wei{-}Ying Ma, Andrew Tomkins, and Xiaodong Zhang, editors, {\em Proceedings of the 17th International Conference on World Wide Web, {WWW} 2008, Beijing, China, April 21-25, 2008}, pages 387--396. {ACM}, 2008.

\bibitem{H2O}
Zhenyu Zhang, Ying Sheng, Tianyi Zhou, Tianlong Chen, Lianmin Zheng, Ruisi Cai, Zhao Song, Yuandong Tian, Christopher R{\'{e}}, Clark~W. Barrett, Zhangyang Wang, and Beidi Chen.
\newblock {H2O:} heavy-hitter oracle for efficient generative inference of large language models.
\newblock In Alice Oh, Tristan Naumann, Amir Globerson, Kate Saenko, Moritz Hardt, and Sergey Levine, editors, {\em Advances in Neural Information Processing Systems 36: Annual Conference on Neural Information Processing Systems 2023, NeurIPS 2023, New Orleans, LA, USA, December 10 - 16, 2023}, 2023.

\bibitem{wildchat}
Wenting Zhao, Xiang Ren, Jack Hessel, Claire Cardie, Yejin Choi, and Yuntian Deng.
\newblock Wildchat: 1m chatgpt interaction logs in the wild.
\newblock In {\em The Twelfth International Conference on Learning Representations, {ICLR} 2024, Vienna, Austria, May 7-11, 2024}. OpenReview.net, 2024.

\bibitem{lmsys}
Lianmin Zheng, Wei{-}Lin Chiang, Ying Sheng, Tianle Li, Siyuan Zhuang, Zhanghao Wu, Yonghao Zhuang, Zhuohan Li, Zi~Lin, Eric~P. Xing, Joseph~E. Gonzalez, Ion Stoica, and Hao Zhang.
\newblock Lmsys-chat-1m: {A} large-scale real-world {LLM} conversation dataset.
\newblock In {\em The Twelfth International Conference on Learning Representations, {ICLR} 2024, Vienna, Austria, May 7-11, 2024}. OpenReview.net, 2024.

\bibitem{sglang}
Lianmin Zheng, Liangsheng Yin, Zhiqiang Xie, Jeff Huang, Chuyue Sun, Cody~Hao Yu, Shiyi Cao, Christos Kozyrakis, Ion Stoica, Joseph~E. Gonzalez, Clark~W. Barrett, and Ying Sheng.
\newblock Efficiently programming large language models using sglang.
\newblock {\em CoRR}, abs/2312.07104, 2023.

\end{thebibliography}

%%%%%%%%%%%%%%%%%%%%%%%%%%%%%%%%%%%%%%%%%%%%%%%%%%%%%%%%%%%%

\clearpage
\appendix
\section{Proof of Theorem 1}
\label{sec:theorem_prove}

\begin{figure}[h]
    \centering
    \includegraphics[width=0.5\linewidth]{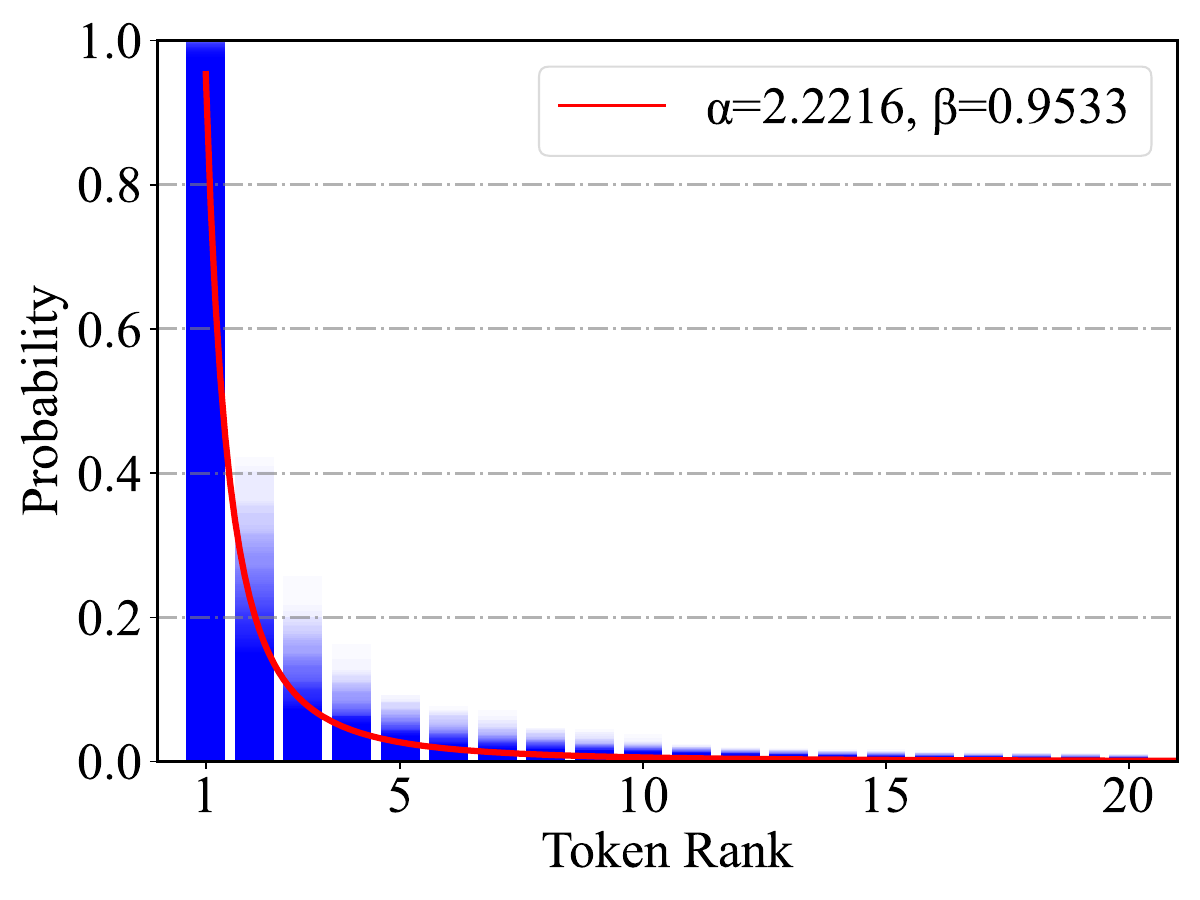}
    \caption{The blue bars represent the probabilities of tokens at different ranks, while the red line shows the power-law function fitted to the observed results. We collected the next-token probability distribution for each token in samples of the ShareGPT dataset using LLaMA3-8B and plotted them cumulatively in a bar chart using a near transparent blue color to reflect the overall distribution pattern. The power-law function is fitted to the average probabilities across token ranks.}
    \label{fig:power_law}
\end{figure}

Figure \ref{fig:power_law} demonstrates the power-law distribution of token probabilities in the LLaMA3-8B model. The proof is presented as follows:

\begin{proof}
Given the the power law distribution probability $P(t_{i, l}) = \beta \, i_l^{-\alpha}$ for each token $t_{i, l}$ ($i$ denoting the rank of the token $t_{i, l}$,  $l$ denoting the position of $t_{i, l}$ in a sequence), the Negative Log-Likelihood (NLL) $\sigma$, and the number of instructions with length $l$ denoted as N$_l$.

For $l = 1$: let $-\log(\beta i^{-\alpha}) \leq \sigma$. This implies:
$$i \leq e^{\frac{\sigma}{\alpha}} \cdot \beta^{\frac{1}{\alpha}} \sim \text{I}_1$$
Thus, N$_1 = \text{I}_1$.

For $l = 2$: $-\log(\beta i_1^{-\alpha}) - \log(\beta i_2^{-\alpha}) \leq \sigma$. This implies:
$$i_2 \leq \frac{e^{\frac{\sigma}{\alpha}} \cdot \beta^{\frac{2}{\alpha}}}{i_1} \sim \text{I}_2$$
Then,
\begin{align*}
  \text{N}_2 &= \sum_{i_1 = 1}^{\text{I}_1} \text{I}_2 \\
         &= e^{\frac{\sigma}{\alpha}} \cdot \beta^{\frac{2}{\alpha}} \sum_{i_1 = 1}^{\text{I}_1} \frac{1}{i_1} \\
         &\approx e^{\frac{\sigma}{\alpha}} \cdot \beta^{\frac{2}{\alpha}} \log \text{I}_1 \\
         &= e^{\frac{\sigma}{\alpha}} \cdot \beta^{\frac{2}{\alpha}} \left(\frac{1}{\alpha}\log \beta + \frac{\sigma}{\alpha}\right)
\end{align*}

For arbitrary $l$: $-\log(\beta i_1^{-\alpha}) - \log(\beta i_2^{-\alpha}) - \cdots - \log(\beta i_l^{-\alpha}) \leq \sigma$. This implies:
$$i_l \leq \frac{e^{\frac{\sigma}{\alpha}} \cdot \beta^{\frac{l}{\alpha}}}{i_1 i_2 \cdots i_{l-1}} \sim \text{I}_l$$
Then,
\begin{align*}
  \text{N}_l &= \sum_{i_1 = 1}^{\text{I}_1} \sum_{i_2 = 1}^{\text{I}_2} \cdots \sum_{i_{l-1} = 1}^{\text{I}_{l-1}} \text{I}_l \\
         &= \sum_{i_1 = 1}^{\text{I}_1} \sum_{i_2 = 1}^{\text{I}_2} \cdots \sum_{i_{l-1} = 1}^{\text{I}_{l-1}} \frac{e^{\frac{\sigma}{\alpha}} \cdot \beta^{\frac{l}{\alpha}}}{i_1 i_2 \cdots i_{l-1}} \\
         &= \sum_{i_1 = 1}^{\text{I}_1} \sum_{i_2 = 1}^{\text{I}_2} \cdots \sum_{i_{l-2} = 1}^{\text{I}_{l-2}} \frac{e^{\frac{\sigma}{\alpha}} \cdot \beta^{\frac{l}{\alpha}}}{i_1 i_2 \cdots i_{l-2}} \sum_{i_{l-1} = 1}^{\text{I}_{l-1}} \frac{1}{i_{l-1}} \\
         &\approx \sum_{i_1 = 1}^{\text{I}_1} \sum_{i_2 = 1}^{\text{I}_2} \cdots \sum_{i_{l-2} = 1}^{\text{I}_{l-2}} \beta^{\frac{1}{\alpha}} \cdot \frac{e^{\frac{\sigma}{\alpha}} \cdot \beta^{\frac{l-1}{\alpha}}}{i_1 i_2 \cdots i_{l-2}} \log \text{I}_{l-1} \\
         &= \sum_{i_1 = 1}^{\text{I}_1} \sum_{i_2 = 1}^{\text{I}_2} \cdots \sum_{i_{l-2} = 1}^{\text{I}_{l-2}} \beta^{\frac{1}{\alpha}} \text{I}_{l-1} \log \text{I}_{l-1}
\end{align*}
Continuing the simplification:
\begin{align*}
  &= \sum_{i_1 = 1}^{\text{I}_1} \sum_{i_2 = 1}^{\text{I}_2} \cdots \sum_{i_{l-3} = 1}^{\text{I}_{l-3}} \beta^{\frac{1}{\alpha}} \sum_{i_{l-2} = 1}^{\text{I}_{l-2}} \frac{e^{\frac{\sigma}{\alpha}} \cdot \beta^{\frac{l-1}{\alpha}}}{i_1 i_2 \cdots i_{l-2}} \log\left(\frac{e^{\frac{\sigma}{\alpha}} \cdot \beta^{\frac{l-1}{\alpha}}}{i_1 i_2 \cdots i_{l-2}}\right) \\
  &= \sum_{i_1 = 1}^{\text{I}_1} \sum_{i_2 = 1}^{\text{I}_2} \cdots \sum_{i_{l-3} = 1}^{\text{I}_{l-3}} \beta^{\frac{2}{\alpha}} \frac{e^{\frac{\sigma}{\alpha}} \cdot \beta^{\frac{l-2}{\alpha}}}{i_1 i_2 \cdots i_{l-3}} \sum_{i_{l-2} = 1}^{\text{I}_{l-2}} \frac{1}{i_{l-2}} \log\left(\frac{e^{\frac{\sigma}{\alpha}} \cdot \beta^{\frac{l-1}{\alpha}}}{i_1 i_2 \cdots i_{l-2}}\right) \\
  &= \sum_{i_1 = 1}^{\text{I}_1} \sum_{i_2 = 1}^{\text{I}_2} \cdots \sum_{i_{l-3} = 1}^{\text{I}_{l-3}} \beta^{\frac{2}{\alpha}} \frac{e^{\frac{\sigma}{\alpha}} \cdot \beta^{\frac{l-2}{\alpha}}}{i_1 i_2 \cdots i_{l-3}} \sum_{i_{l-2} = 1}^{\text{I}_{l-2}} \frac{1}{i_{l-2}} \left(\frac{1}{\alpha} \log \beta + \log\left(\frac{e^{\frac{\sigma}{\alpha}} \cdot \beta^{\frac{l-2}{\alpha}}}{i_1 i_2 \cdots i_{l-3}}\right) - \log i_{l-2}\right) \\
  &= \sum_{i_1 = 1}^{\text{I}_1} \sum_{i_2 = 1}^{\text{I}_2} \cdots \sum_{i_{l-3} = 1}^{\text{I}_{l-3}} \beta^{\frac{2}{\alpha}} \text{I}_{l-2} \sum_{i_{l-2} = 1}^{\text{I}_{l-2}} \frac{1}{i_{l-2}} \left(\frac{1}{\alpha} \log \beta + \log \text{I}_{l-2} - \log i_{l-2}\right) \\
  &\approx \sum_{i_1 = 1}^{\text{I}_1} \sum_{i_2 = 1}^{\text{I}_2} \cdots \sum_{i_{l-3} = 1}^{\text{I}_{l-3}} \beta^{\frac{2}{\alpha}} \text{I}_{l-2} \left[\frac{1}{\alpha} \log \beta \cdot \log \text{I}_{l-2} + (\log \text{I}_{l-2})^2 - \frac{1}{2}(\log \text{I}_{l-2})^2\right] \\
  &= \sum_{i_1 = 1}^{\text{I}_1} \sum_{i_2 = 1}^{\text{I}_2} \cdots \sum_{i_{l-3} = 1}^{\text{I}_{l-3}} \beta^{\frac{2}{\alpha}} \text{I}_{l-2} \left[\frac{1}{\alpha} \log \beta \cdot \log \text{I}_{l-2} + \frac{1}{2}(\log \text{I}_{l-2})^2\right] \\
  &= \sum_{i_1 = 1}^{\text{I}_1} \sum_{i_2 = 1}^{\text{I}_2} \cdots \sum_{i_{l-3} = 1}^{\text{I}_{l-3}} \beta^{\frac{2}{\alpha}} \text{I}_{l-2} \cdot \log \text{I}_{l-2} \cdot \left[\frac{1}{2} (\log \text{I}_{l-2} + 2 \frac{1}{\alpha} \log \beta)\right]
\end{align*}

We continuing the simplification and skip the cumbersome process:
$$= \sum_{i_1 = 1}^{\text{I}_1} \sum_{i_2 = 1}^{\text{I}_2} \cdots \sum_{i_{l-4} = 1}^{\text{I}_{l-4}} \beta^{\frac{3}{\alpha}} \text{I}_{l-3} \cdot \log \text{I}_{l-3} \cdot \left[\frac{1}{6} (\log \text{I}_{l-3} + 3 \frac{1}{\alpha} \log \beta)^2\right]$$

$$= \sum_{i_1 = 1}^{\text{I}_1} \sum_{i_2 = 1}^{\text{I}_2} \cdots \sum_{i_{l-5} = 1}^{\text{I}_{l-5}} \beta^{\frac{4}{\alpha}} \text{I}_{l-4} \cdot \log \text{I}_{l-4} \cdot \left[\frac{1}{24} (\log \text{I}_{l-4} + 4 \frac{1}{\alpha} \log \beta)^3\right]$$

Finally, we can obtain the result by eliminating all summation symbols:
\begin{align*}
  &= \beta^{\frac{l-1}{\alpha}} \text{I}_1 \log \text{I}_1 \cdot \frac{1}{(l-1)!}[\log \text{I}_1 + (l-1)\frac{1}{\alpha}\log \beta]^{l-2} \\
  &= \frac{\beta^{\frac{l}{\alpha}} \cdot e^{\frac{\sigma}{\alpha}}}{(l-1)!} \left(\frac{1}{\alpha} \log \beta + \frac{\sigma}{\alpha}\right) \left(\frac{l}{\alpha} \log \beta + \frac{\sigma}{\alpha}\right)^{l-2}
\end{align*}

Since the $\beta$ closes to 1, we further simplify the equation as:

\begin{align*}
    N_L \approx {e^{\frac{\sigma}{\alpha}} \cdot \left(\frac{\sigma}{\alpha}\right)^{L-1}}/{(L-1)!}
\end{align*}

\end{proof}

\section{Incorrect cache hits in GPTCache}
\label{GPTCache}

We employ the provided Albert-small\citep{sbert} embedding model for GPTCache. Given that the Albert-small embedding model exclusively supports English, our evaluations are limited to the English samples within the WildChat dataset. We conduct experiments with similarity thresholds of 0.9 and 0.95, for which GPTCache reports claimed hit rates of 22.00\% and 17.68\%, respectively. To assess the semantic accuracy of the cache hits, we use the DeepSeek V3 model with the prompt of \emph{"Please determine whether following two queries are similar (Yes or No). Similar means they will have similar answers. Remember that you don't need to answer the question and you should not give any other additional response \textbackslash nQuery1: \{text1\}\textbackslash nQuery2: \{text2\}"}. 

Based on this evaluation, we found that the mismatching rates are 29.86\% and 33.55\% for the thresholds of 0.9 and 0.95, respectively. Table \ref{tab:text_examples} provides examples of both matched and mismatched instruction pairs. These results suggest that semantic equivalence between an instruction identified as a cache hit by GPTCache and the user's actual query cannot be reliably guaranteed. As a result, misleading responses could potentially be returned to users making queries. This highlights the practical challenges of deploying GPTCache in enterprise-level applications.

\begin{table}[htbp]
    \centering
    \caption{Example instruction pairs identified as hits by GPTCache and the corresponding judgments by DeepSeek V3}
    \label{tab:text_examples}
    \resizebox{\columnwidth}{!}{
        \begin{tabular}{p{0.07\textwidth} p{0.47\textwidth} p{0.43\textwidth}}
            \toprule
            DSV3 & User Instruction & GPTCache Hit Instruction \\
            \midrule
            {[NO]} & what's a "kinky date"?  & what's the date today? \\
            {[NO]} & create an image of the moon  & generate image of moon lovers \\
            {[NO]} & what if jon snow was batman?  & what if batman met a snow man? \\
            {[NO]} & what is alan turing's views on consciousness?  & what were hugh everett's views on consciousness? \\
            {[NO]} & make an outline for this blog post: escrow vs fe  & make a blog post: escrow vs fe \\
            {[NO]} & is water wet?  & write me a syllogism that proves that water is wet \\
            {[NO]} & now, tell me the latest news of february 2024!  & current news in the world as of 2024 \\
            {[NO]} & multi-modality image fusion research paper  & explain multi-modality fusion in simple terms \\
            {[NO]} & create a calculator by using python and then javascript  & create a python code for calculator \\
            {[NO]} & how large does a planet have to be to be considered a planet?  & how big is the earth? \\
            {[NO]} & why should we study internation jusitce issues  & why we study \\
            {[NO]} & how to create blog post  & how can you help me on creating my blog site \\
            {[NO]} & can you write me a blog post about it's best for you to do your child's homework like this?  & can you give me a blog post about supporting your child’s natural curiosity and exploration \\
            \midrule
            {[YES]} & what's the day today?  & what day is it today? \\
            {[YES]} & best method to lose weight  & how to lose weight \\
            {[YES]} & why are slavic women beautiful?  & why are slavic girls pretty? \\
            {[YES]} & are you capable of analyzing images ?  & are you able to analyze images \\
            {[YES]} & what are some popular chinese apps right now  & what are some popular apps in china? \\
            \bottomrule
        \end{tabular}
    }
\end{table}

\section{Effect of Data Scale}
\label{sec:data_scale}

\begin{figure}[h]
    \centering
    \subcaptionbox{Effect of Train Ratio on the WildChat dataset.\label{fig:train_ratio}}
    {\includegraphics[width=0.49\linewidth]{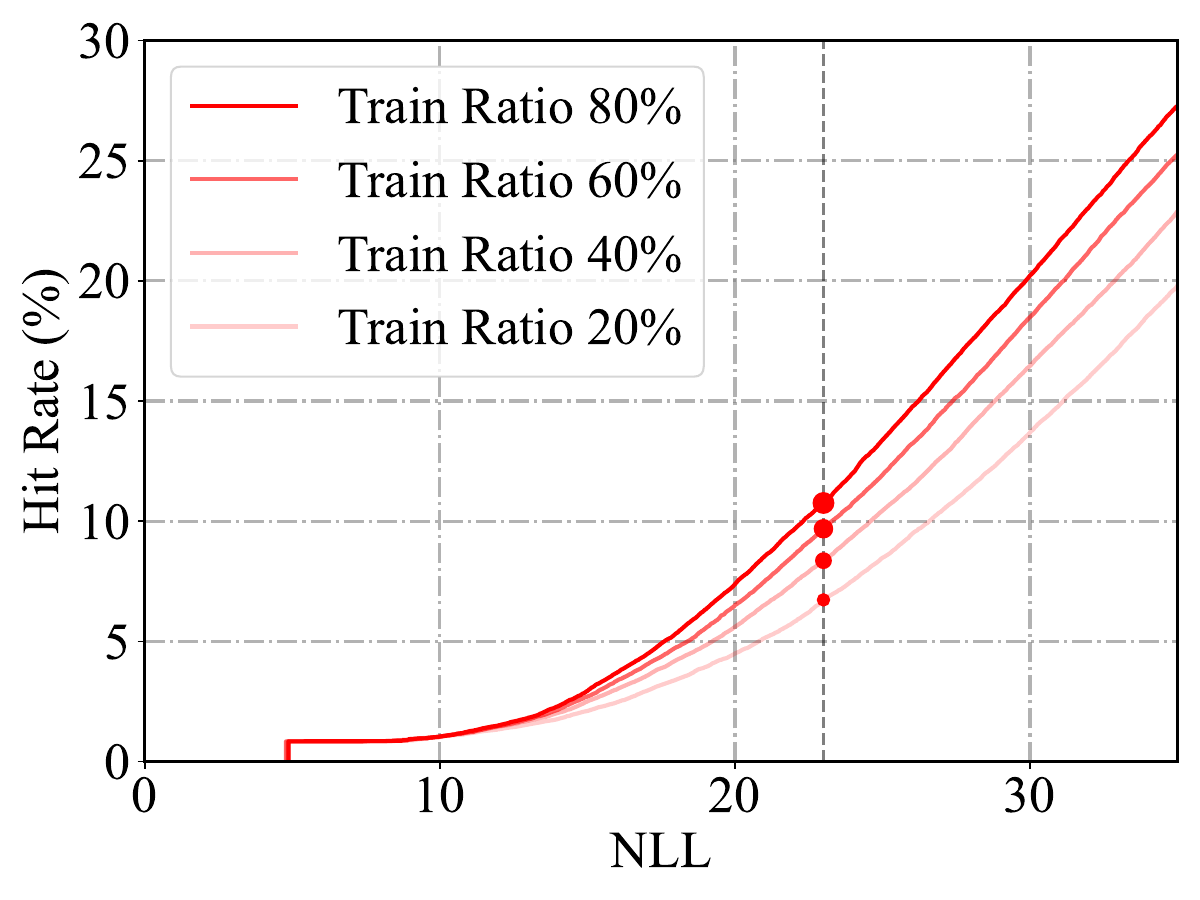}}
    \hfill
    \subcaptionbox{Effect of Test Ratio across different datasets.\label{fig:test_ratio}}
    {\includegraphics[width=0.49\linewidth]{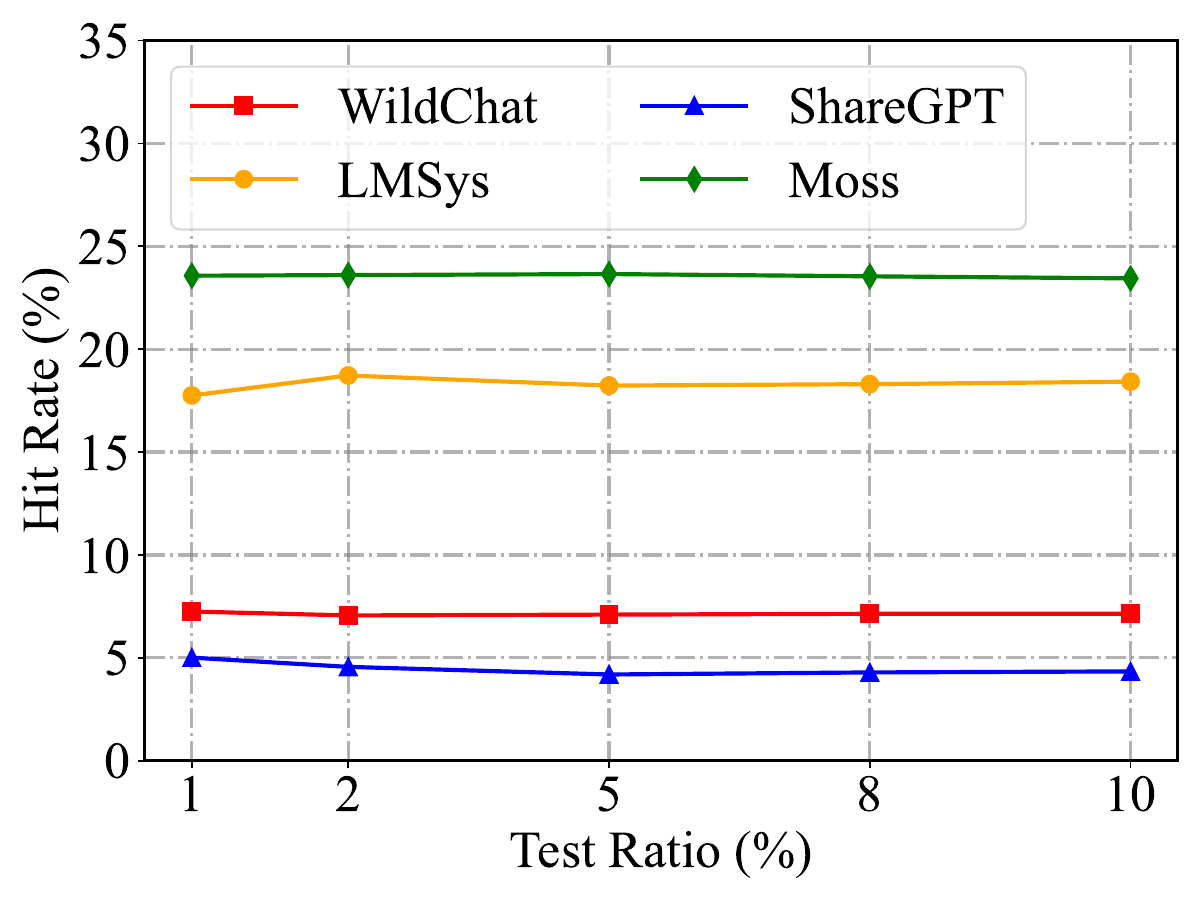}}
    \caption{Evaluation of the impact of train and test ratio.}
    \label{fig:ratio}
\end{figure}

As depicted in Figure \ref{fig:train_ratio}, the performance of InstCache progressively improves with larger training data sizes, indicating that access to a larger set of observed instructions can better unleashed the potential of our approach. For large model service providers, with significantly more serving data, an even higher level of performance than reported in this paper can be expected.

To evaluate the impact of test set size, we conducted hit rate evaluation experiments using 1\%, 2.5\%, 5\%, and 7.5\% of the test data to replace the original 10\% test set. As shown in Figure \ref{fig:test_ratio}, varying the size of the test set has minimal effect on the hit rates of InstCache.

\section{Cache Details}
\label{sec:cache_details}

\begin{table}[htbp]
    \centering
    \caption{Cache Details. The pre-population time is evaluated under a distributed setup using three A100 GPUs.}
    \resizebox{\columnwidth}{!}{
        \begin{tabular}{c|c|c|c|c|c|c|c}
            \toprule
            \textbf{Datasets} & $\sigma$ & \textbf{Hit Rate} & \textbf{Instructions Num} & \textbf{Avg Response Length} & \textbf{Mem Size (GB)} & \textbf{Storage Size (GB)} & \textbf{Pre-population Time (h)} \\
            \hline
            \multirow{7}{*}{\begin{tabular}[c]{@{}c@{}} WildChat \end{tabular}} & 15 & 2.68\% & 34,440 & 373 & 0.25 & 0.08 & 0.05 \\
             & 16 & 3.34\% & 97,641 & 376 & 0.71 & 0.23 & 0.08 \\
             & 17 & 4.13\% & 276,170 & 375 & 2.00 & 0.64 & 0.16 \\
             & 18 & 5.03\% & 788,387 & 373 & 5.69 & 1.83 & 0.37 \\
             & 19 & 6.02\% & 2,269,127 & 381 & 16.68 & 5.36 & 1.33 \\
             & 20 & 7.14\% & 6,531,272 & 380 & 47.90 & 15.35 & 2.45 \\
             & 21 & 8.21\% & 18,698,939 & 392 & 141.12 & 45.22 & 9.85 \\
            \hline
            \multirow{7}{*}{\begin{tabular}[c]{@{}c@{}} LMSys \end{tabular}} & 15 & 9.64\% & 97,910 & 455 & 0.86 & 0.27 & 0.05 \\
             & 16 & 11.60\% & 228,615 & 423 & 1.85 & 0.59 & 0.08 \\
             & 17 & 13.20\% & 584,026 & 392 & 4.36 & 1.40 & 0.16 \\
             & 18 & 15.07\% & 1,559,410 & 389 & 11.55 & 3.69 & 0.36 \\
             & 19 & 16.77\% & 4,091,897 & 381 & 30.31 & 9.71 & 0.91 \\
             & 20 & 18.42\% & 11,254,940 & 379 & 80.81 & 25.90 & 2.17 \\
             & 21 & 20.12\% & 30,183,132 & 381 & 217.88 & 69.95 & 5.55 \\
            \hline
            \multirow{7}{*}{\begin{tabular}[c]{@{}c@{}} ShareGPT \end{tabular}} & 15 & 1.34\% & 22,657 & 245 & 0.10 & 0.03 & 0.04 \\
             & 16 & 1.85\% & 69,164 & 271 & 0.35 & 0.11 & 0.06 \\
             & 17 & 2.33\% & 207,843 & 277 & 1.09 & 0.35 & 0.13 \\
             & 18 & 2.90\% & 620,859 & 305 & 3.60 & 1.15 & 0.29 \\
             & 19 & 3.64\% & 1,831,202 & 312 & 10.87 & 3.47 & 0.60 \\
             & 20 & 4.34\% & 5,344,636 & 331 & 33.63 & 10.75 & 1.60 \\
             & 21 & 4.95\% & 14,713,916 & 339 & 95.38 & 30.32 & 8.00\\
            \hline
            \multirow{7}{*}{\begin{tabular}[c]{@{}c@{}} Moss \end{tabular}} & 15 & 0.72\% & 43,908 & 416 & 0.35 & 0.11 & 0.05 \\
             & 16 & 2.80\% & 250,393 & 445 & 2.20 & 0.70 & 0.08 \\
             & 17 & 6.90\% & 994,008 & 469 & 9.00 & 2.86 & 0.19 \\
             & 18 & 12.33\% & 3,286,293 & 468 & 29.74 & 9.45 & 0.43 \\
             & 19 & 18.00\% & 9,833,299 & 472 & 89.62 & 28.54 & 1.12 \\
             & 20 & 23.44\% & 27,645,188 & 467 & 249.60 & 79.42 & 2.97 \\
            \bottomrule
        \end{tabular}
    }
    \label{tab:my_label}
\end{table}

% 考虑放样例

\begin{table}[t]
    \centering
    \caption{InstCache hit examples on test set. The instructions shown in the table do not appear in the training set}
    \label{tab:cache_hit_not_in_train}
    \begin{tabular}{p{0.1\textwidth}p{0.8\textwidth}}
        \toprule
        \textbf{Datasets} & \textbf{InstCache Hit Examples} \\
        \midrule
        \multirow{11}{*}{\begin{tabular}[c]{@{}c@{}} WildChat \end{tabular}} & what is aigc? \\
                          & teach me french \\
                          & what is personality development? \\
                          & what is security technology management? \\
                          & explain quantum computing \\
                          & create a python discord bot \\
                          & what is method statement? \\
                          & what is spiritual science? \\
                          & make a spongebob episode \\
                          & what is the difference between bayesian and frequentist statistics? \\
        \midrule
        \multirow{11}{*}{\begin{tabular}[c]{@{}c@{}} LMsys \end{tabular}}    &  write me a webserver in rust \\
                          & how can i improve my sleep? \\
                          & what is cumulative distribution function \\
                          & large language models for data science \\
                          & write a program to compute the fibonacci sequence \\
                          & what are common tech interview questions \\
                          & the largest city in germany \\
                          & explain the main features of the rust programming language \\
                          & what is self-attention in transformers? \\
                          & descriptive answer for matplotlib log in python with proper code example and outputs. \\
        \midrule
        \multirow{11}{*}{\begin{tabular}[c]{@{}c@{}} ShareGPT \end{tabular}}  &  who is elon musk? \\
                          & what is prompt engineering? \\
                          & show me how to build a database \\
                          & how to use chatgpt api \\
                          & what is the chemical symbol for hydrogen? \\
                          & what are the advantages of acrylic painting over oil painting? \\
                          & write a basic neural network in c++ \\
                          & is python the best language for machine learning? \\
                          & write a rap about the status of chatgpt. \\
                          & what to do for 10 days in florence? \\
                          & what is the difference between deep learning, machine learning, and artificial intelligence? \\
        \midrule
        \multirow{12}{*}{\begin{tabular}[c]{@{}c@{}} Moss \end{tabular}}     & write a short story about the impact of climate change. \\
                          & design a poster highlighting the benefits and risks of using ai in everyday life. \\
                          & what strategies can i use for brainstorming new ideas? \\
                          & how do i optimize the user experience on my website? \\
                          & design a website that explains how ai can be used to solve everyday problems \\
                          & design a function in javascript that takes two numbers as input and returns their sum \\
                          & write a function to find the maximum depth of a binary tree in c++ \\
                          & what resources do you recommend for learning coding fundamentals? \\
                          &  create a python script to generate random numbers from 1-100 \\
                          &  what is the best way to create an effective budget? \\
                          &  explain how to use pandas dataframes for data manipulation. \\
                          &  generate an essay exploring the ethical implications of using artificial intelligence in decision-making processes. \\
                          
        \bottomrule
    \end{tabular}
\end{table}

%%%%%%%%%%%%%%%%%%%%%%%%%%%%%%%%%%%%%%%%%%%%%%%%%%%%%%%%%%%%

\clearpage
\section*{NeurIPS Paper Checklist}

\begin{enumerate}

\item {\bf Claims}
    \item[] Question: Do the main claims made in the abstract and introduction accurately reflect the paper's contributions and scope?
    \item[] Answer: \answerYes{} % Replace by \answerYes{}, \answerNo{}, or \answerNA{}.
    \item[] Justification: The main claims in the abstract and introduction accurately reflect the paper’s contributions and scope.

\item {\bf Limitations}
    \item[] Question: Does the paper discuss the limitations of the work performed by the authors?
    \item[] Answer: \answerYes{} % Replace by \answerYes{}, \answerNo{}, or \answerNA{}.
    \item[] Justification: We discuss the limitations in the Discussion section.

\item {\bf Theory assumptions and proofs}
    \item[] Question: For each theoretical result, does the paper provide the full set of assumptions and a complete (and correct) proof?
    \item[] Answer: \answerYes{} % Replace by \answerYes{}, \answerNo{}, or \answerNA{}.
    \item[] Justification: The paper provides a complete set of assumptions and includes correct and rigorous proof for the theoretical result in Section \ref{sec:instcache} and Appendix \ref{sec:theorem_prove}.

    \item {\bf Experimental result reproducibility}
    \item[] Question: Does the paper fully disclose all the information needed to reproduce the main experimental results of the paper to the extent that it affects the main claims and/or conclusions of the paper (regardless of whether the code and data are provided or not)?
    \item[] Answer: \answerYes{} % Replace by \answerYes{}, \answerNo{}, or \answerNA{}.
    \item[] Justification: We provide the detailed algorithm in the Section \ref{sec:implementation} and present the reproduction details in Section \ref{sec:exper_setup}.

\item {\bf Open access to data and code}
    \item[] Question: Does the paper provide open access to the data and code, with sufficient instructions to faithfully reproduce the main experimental results, as described in supplemental material?
    \item[] Answer: \answerYes{} % Replace by \answerYes{}, \answerNo{}, or \answerNA{}.
    \item[] Justification: All datasets used in this paper are publicly available on Hugging Face and we provide the code with anonymous url.

\item {\bf Experimental setting/details}
    \item[] Question: Does the paper specify all the training and test details (e.g., data splits, hyperparameters, how they were chosen, type of optimizer, etc.) necessary to understand the results?
    \item[] Answer: \answerYes{} % Replace by \answerYes{}, \answerNo{}, or \answerNA{}.
    \item[] Justification: We present the experimental details in Section \ref{sec:exper_setup}

\item {\bf Experiment statistical significance}
    \item[] Question: Does the paper report error bars suitably and correctly defined or other appropriate information about the statistical significance of the experiments?
    \item[] Answer: \answerNo{} % Replace by \answerYes{}, \answerNo{}, or \answerNA{}.
    \item[] Justification: Our experiments are minimally affected by randomness, and the results are consistently reproducible. This experimental setup is also consistent with existing studies\citep{vllm, sglang}.

\item {\bf Experiments compute resources}
    \item[] Question: For each experiment, does the paper provide sufficient information on the computer resources (type of compute workers, memory, time of execution) needed to reproduce the experiments?
    \item[] Answer: \answerYes{} % Replace by \answerYes{}, \answerNo{}, or \answerNA{}.
    \item[] Justification: We provide detailed information about our compute resources in Section \ref{sec:exper_setup}.
    
\item {\bf Code of ethics}
    \item[] Question: Does the research conducted in the paper conform, in every respect, with the NeurIPS Code of Ethics \url{https://neurips.cc/public/EthicsGuidelines}?
    \item[] Answer: \answerYes{} % Replace by \answerYes{}, \answerNo{}, or \answerNA{}.
    \item[] Justification: The research conducted in the paper fully conforms to the NeurIPS Code of Ethics in all respects.

\item {\bf Broader impacts}
    \item[] Question: Does the paper discuss both potential positive societal impacts and negative societal impacts of the work performed?
    \item[] Answer: \answerYes{} % Replace by \answerYes{}, \answerNo{}, or \answerNA{}.
    \item[] Justification: We discuss the societal impacts in the Discussion Section.
    
\item {\bf Safeguards}
    \item[] Question: Does the paper describe safeguards that have been put in place for responsible release of data or models that have a high risk for misuse (e.g., pretrained language models, image generators, or scraped datasets)?
    \item[] Answer: \answerNA{} % Replace by \answerYes{}, \answerNo{}, or \answerNA{}.
    \item[] Justification: This paper does not release any new dataset or model.

\item {\bf Licenses for existing assets}
    \item[] Question: Are the creators or original owners of assets (e.g., code, data, models), used in the paper, properly credited and are the license and terms of use explicitly mentioned and properly respected?
    \item[] Answer: \answerYes{} % Replace by \answerYes{}, \answerNo{}, or \answerNA{}.
    \item[] Justification: We properly cite all the papers that produced the datasets and models used in this work.

\item {\bf New assets}
    \item[] Question: Are new assets introduced in the paper well documented and is the documentation provided alongside the assets?
    \item[] Answer: \answerNA{} % Replace by \answerYes{}, \answerNo{}, or \answerNA{}.
    \item[] Justification: The paper does not release new assets.

\item {\bf Crowdsourcing and research with human subjects}
    \item[] Question: For crowdsourcing experiments and research with human subjects, does the paper include the full text of instructions given to participants and screenshots, if applicable, as well as details about compensation (if any)? 
    \item[] Answer: \answerNA{} % Replace by \answerYes{}, \answerNo{}, or \answerNA{}.
    \item[] Justification: The paper does not involve crowdsourcing nor research with human subjects.

\item {\bf Institutional review board (IRB) approvals or equivalent for research with human subjects}
    \item[] Question: Does the paper describe potential risks incurred by study participants, whether such risks were disclosed to the subjects, and whether Institutional Review Board (IRB) approvals (or an equivalent approval/review based on the requirements of your country or institution) were obtained?
    \item[] Answer: \answerNA{} % Replace by \answerYes{}, \answerNo{}, or \answerNA{}.
    \item[] Justification: The paper does not involve crowdsourcing nor research with human subjects.

\item {\bf Declaration of LLM usage}
    \item[] Question: Does the paper describe the usage of LLMs if it is an important, original, or non-standard component of the core methods in this research? Note that if the LLM is used only for writing, editing, or formatting purposes and does not impact the core methodology, scientific rigorousness, or originality of the research, declaration is not required.
    %this research? 
    \item[] Answer: \answerNA{} % Replace by \answerYes{}, \answerNo{}, or \answerNA{}.
    \item[] Justification: The core method development in this research does not involve LLMs as any important, original, or non-standard components.

\end{enumerate}

\end{document}